\documentclass[10pt,twocolumn,letterpaper]{article}

\usepackage[pagenumbers]{cvpr}

\usepackage{graphicx}
\usepackage{booktabs}
\usepackage{algorithm}
\usepackage{algpseudocode}
\usepackage{multirow}
\usepackage{makecell}
\usepackage[table]{xcolor}
\usepackage{arydshln}
\usepackage{amsmath}
\usepackage{amssymb}
\usepackage{caption}
\usepackage{wrapfig}
\usepackage{tcolorbox}

\definecolor{cvprblue}{rgb}{0.21,0.49,0.74}
\usepackage[pagebackref,breaklinks,colorlinks,allcolors=cvprblue]{hyperref}

\def\paperID{*****}
\def\confName{CVPR}
\def\confYear{2026}

\title{MemRoPE: Training-Free Infinite Video Generation \\ via Evolving Memory Tokens}

\author{
Youngrae Kim$^{*}$ \quad
Qixin Hu$^{*}$ \quad
C.-C. Jay Kuo \quad
Peter A.  Beerel\\
University of Southern California
}

\begin{document}

% \maketitle
\twocolumn[{
  \maketitle
  \begin{center}
    \vspace{-10pt}
    Project Page: \url{https://memrope.github.io}
    \vspace{10pt}
    
    \includegraphics[width=\linewidth]{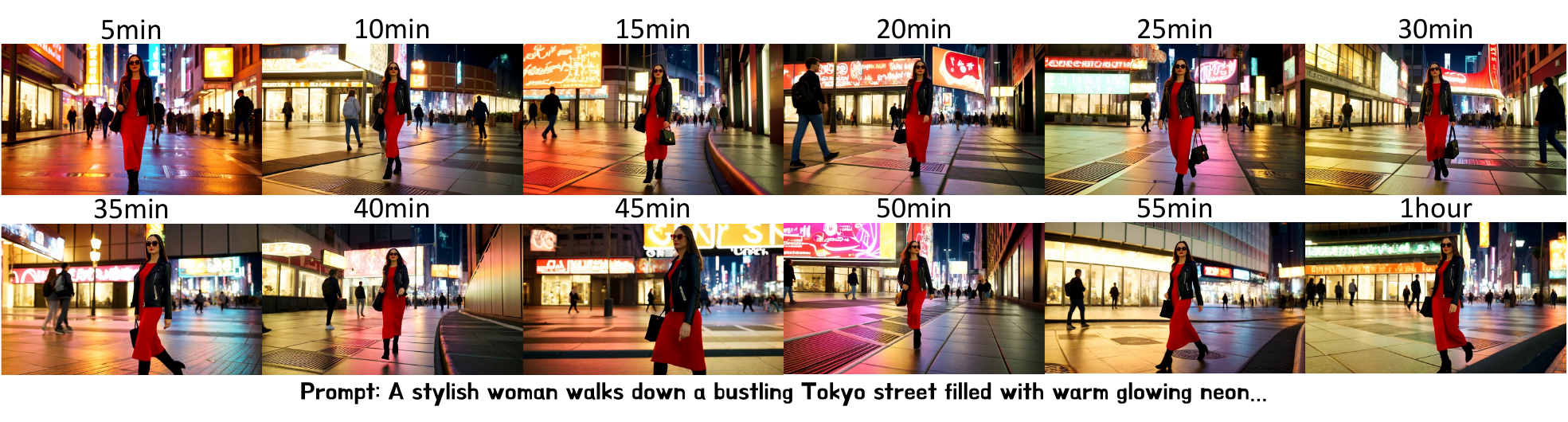}
    \captionof{figure}{\textbf{Training-free unbounded video generation.} Our MemRoPE requires \textbf{\textit{no additional training}} and enables \textbf{\textit{unlimited generation}} with a fixed-size KV cache. We demonstrate a continuous one-hour generation process that perfectly preserves both subject identity and visual fidelity throughout.}
    \label{fig:one_hour}
  \end{center}
}]

% \let\thefootnote\relax\footnotetext{${}^{*}$Equal contribution.}
% \maketitle
\renewcommand{\thefootnote}{}
\footnotetext{\hspace{-1.5em}${}^{*}$Equal contribution.}
\renewcommand{\thefootnote}{\arabic{footnote}}

\begin{abstract}
Autoregressive diffusion enables real-time frame streaming, yet existing sliding-window caches discard past context, causing fidelity degradation, identity drift, and motion stagnation over long horizons. Current approaches preserve a fixed set of early tokens as attention sinks, but this static anchor cannot reflect the evolving content of a growing video. We introduce MemRoPE, a training-free framework with two co-designed components. Memory Tokens continuously compress all past keys into dual long-term and short-term streams via exponential moving averages, maintaining both global identity and recent dynamics within a fixed-size cache. Online RoPE Indexing caches unrotated keys and applies positional embeddings dynamically at attention time, ensuring the aggregation is free of conflicting positional phases. These two mechanisms are mutually enabling: positional decoupling makes temporal aggregation well-defined, while aggregation makes fixed-size caching viable for unbounded generation. Extensive experiments validate that MemRoPE outperforms existing methods in temporal coherence, visual fidelity, and subject consistency across minute- to hour-scale generation.
\end{abstract}

\begin{figure*}[t]
\centering
\includegraphics[width=0.9\textwidth]{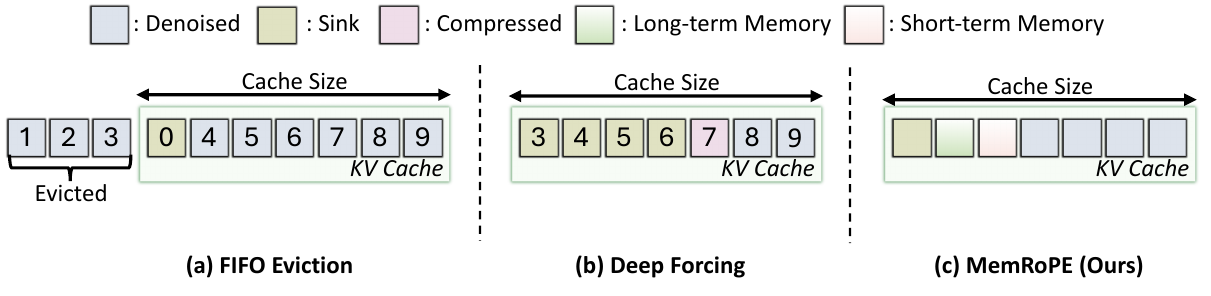}
\caption{\textbf{KV cache structures for long video generation.}
\textbf{(a)}~FIFO eviction~\cite{selfforcing,causvid,longlive,infinityrope} maintains an initial sink frame while discarding the oldest remaining frames when the cache is full, losing distant context.
\textbf{(b)}~Deep Forcing~\cite{deepforcing} dedicates over half the cache to static sink tokens and selects a small number of compressed tokens via attention-based importance scoring, which often causes temporal instability in the generated sequence (see \cref{fig:pc_failure}).
\textbf{(c)}~MemRoPE preserves a sink frame and manages distinct \textbf{long-} and \textbf{short-term} \textbf{memories} (\cref{sec:memory}). By storing all keys \textbf{without RoPE}, it prevents positional interference from corrupting the stored features, thereby enabling stable  memory aggregation (\cref{sec:online_rope}).}
\label{fig:method_comparison}
\end{figure*}

\section{Introduction}
\label{sec:intro}

% Long-form video generation from a single prompt is essential for applications such as continuous world simulation~\cite{feng2024matrix,hong2025relic,brooks2024video,sun2025worldplay,yang2026stableworld}, cinematic long takes~\cite{moviegen,elmoghany2025survey}, and synthetic data generation~\cite{hu2023gaia,ren2025cosmos}, where temporal coherence must be maintained over minutes to hours.
% Recent video diffusion models~\cite{liu2024sora,yang2024cogvideox,kong2024hunyuanvideo,wan} produce cinematic-quality clips, but are limited to fixed lengths in a single forward pass; extending them to arbitrarily long videos requires a fundamentally different generation paradigm.

% Introducing why we need long videos
Recent video diffusion models~\cite{liu2024sora,yang2024cogvideox,kong2024hunyuanvideo,wan} excel at producing cinematic-quality clips, but are inherently limited to fixed lengths in a single forward pass. Beyond mere short-clip synthesis, the broader objective is to simulate evolving visual worlds that maintain persistent identity and temporal coherence over minutes to hours. This long-form generation is essential for powering advanced applications such as continuous world simulation~\cite{feng2024matrix,hong2025relic,brooks2024video,sun2025worldplay,yang2026stableworld}, cinematic long takes~\cite{moviegen,elmoghany2025survey}, and synthetic data generation~\cite{hu2023gaia,ren2025cosmos}. Extending them to achieve arbitrarily long, coherent videos requires a fundamentally different generation paradigm.

% Introducing the AR and its drawbacks
Autoregressive video diffusion generates frames sequentially from a pretrained model, naturally enabling variable-length generation.
CausVid~\cite{causvid} distills a bidirectional DiT into a causal generator via DMD~\cite{dmd} for real-time synthesis, and Self Forcing~\cite{selfforcing} closes the train-inference gap by conditioning on self-generated frames. LongLive~\cite{longlive} introduces streaming long tuning that enables long video training to align training and inference. While these models produce high-quality frames in real-time, they are limited to a finite temporal horizon. $\infty$-RoPE~\cite{infinityrope,rope} resolves the resulting positional extrapolation by block-relative RoPE, enabling generation beyond the 1024-frame limit.
However, FIFO eviction (\cref{fig:method_comparison}(a)) in the sliding-window KV cache still discards past context as generation proceeds, leading to progressive error accumulation~\cite{selfforcing,longlive,deepforcing,infinityrope}.

\begin{figure*}[t]
\centering
\begin{subfigure}[t]{0.24\textwidth}
    \centering
    \includegraphics[width=\textwidth]{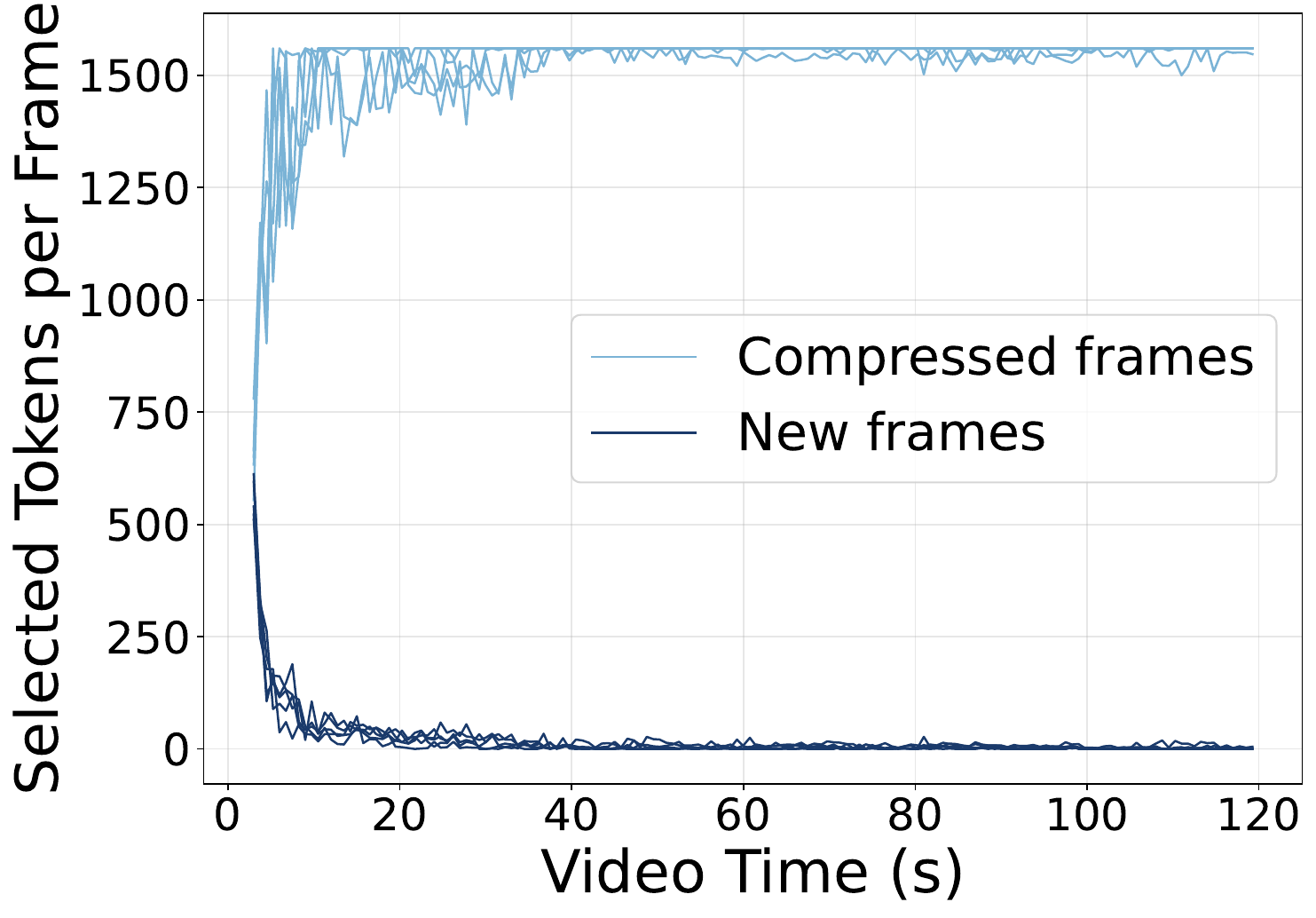}
    \caption{Token selection}
    \label{fig:pc_token}
\end{subfigure}
\hfill
\begin{subfigure}[t]{0.24\textwidth}
    \centering
    \includegraphics[width=\textwidth]{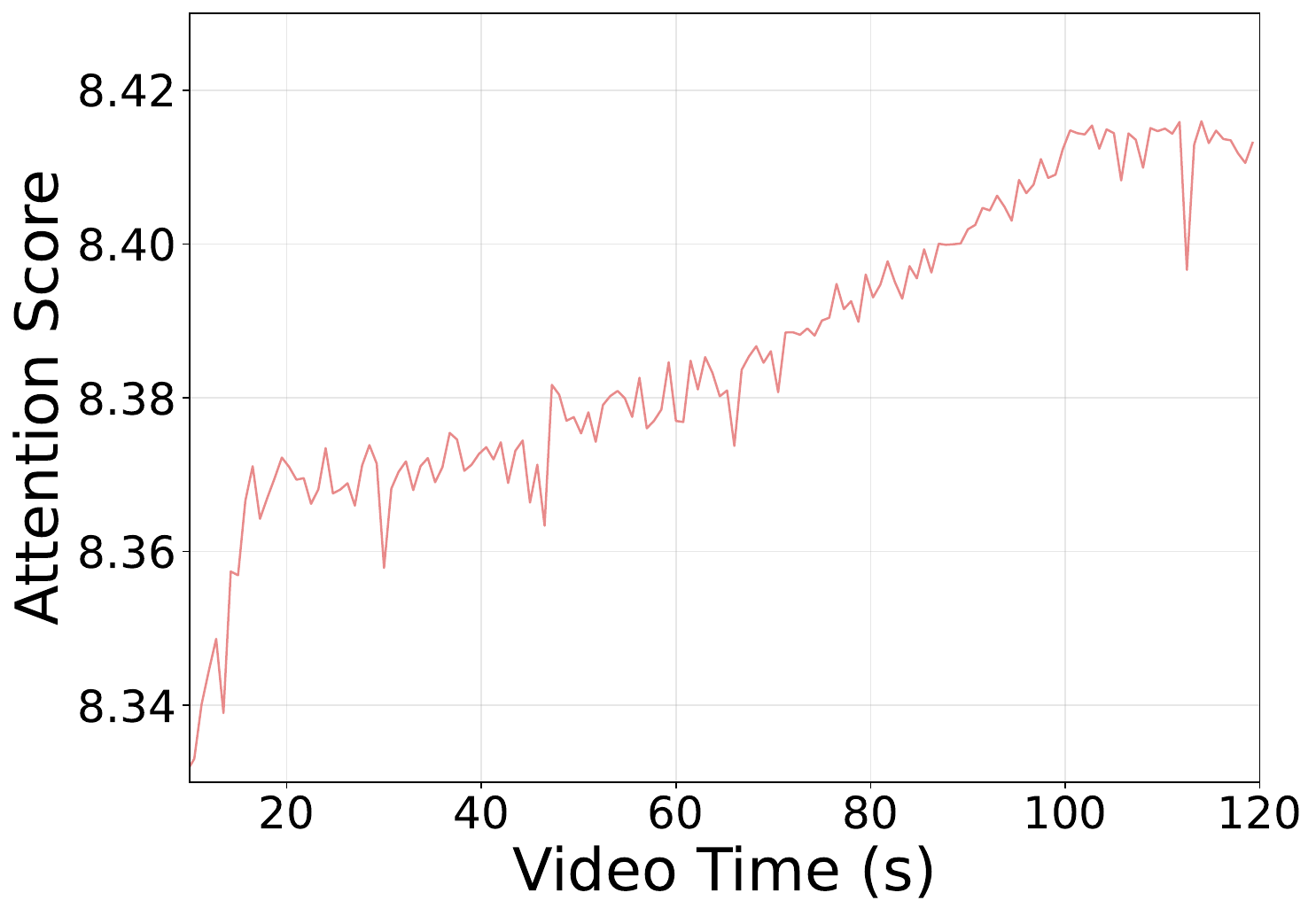}
    \caption{Attention score}
    \label{fig:pc_attn}
\end{subfigure}
\hfill
\begin{subfigure}[t]{0.24\textwidth}
    \centering
    \includegraphics[width=\textwidth]{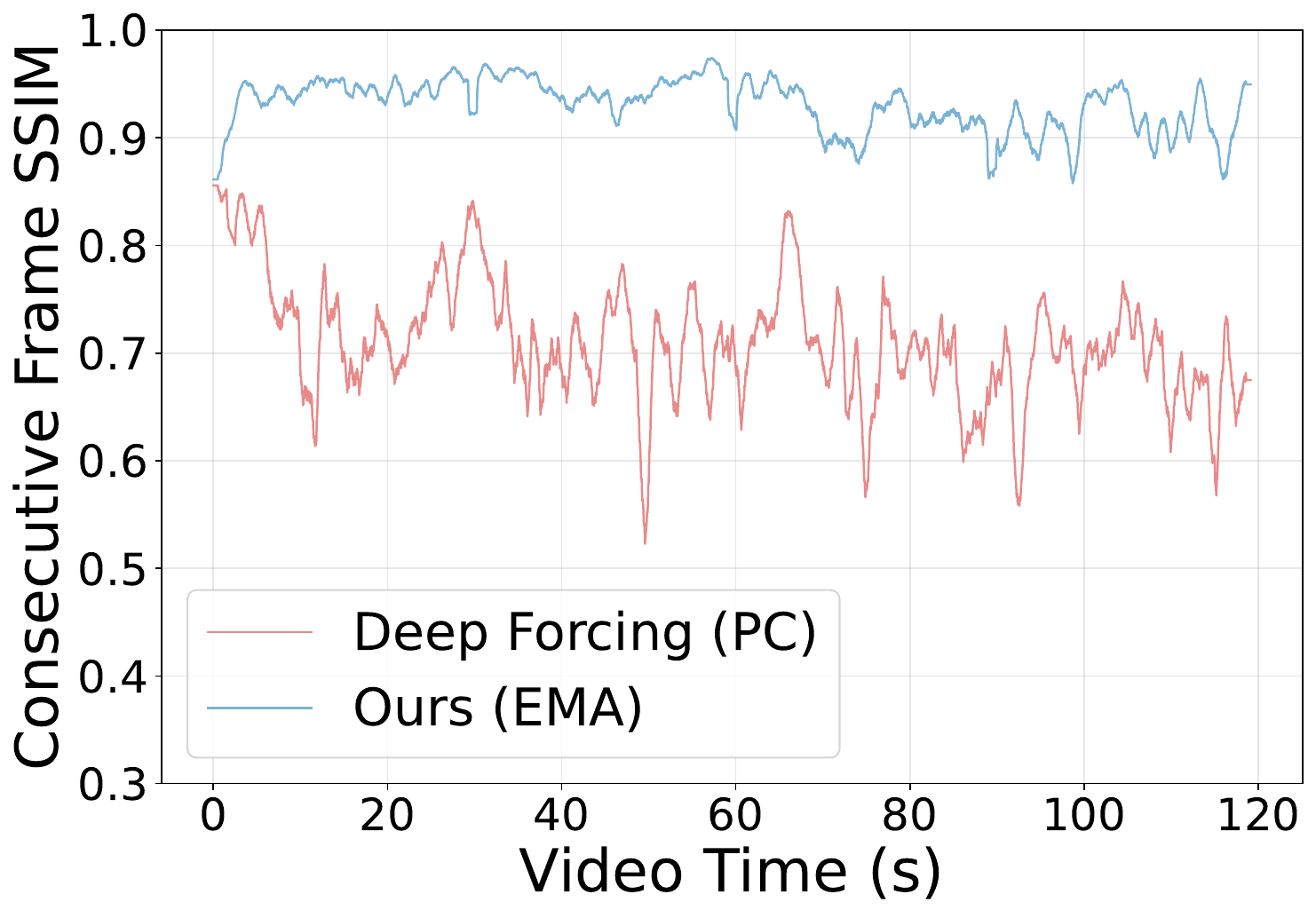}
    \caption{Consecutive SSIM}
    \label{fig:pc_ssim}
\end{subfigure}
\hfill
\begin{subfigure}[t]{0.24\textwidth}
    \centering
    \includegraphics[width=\textwidth]{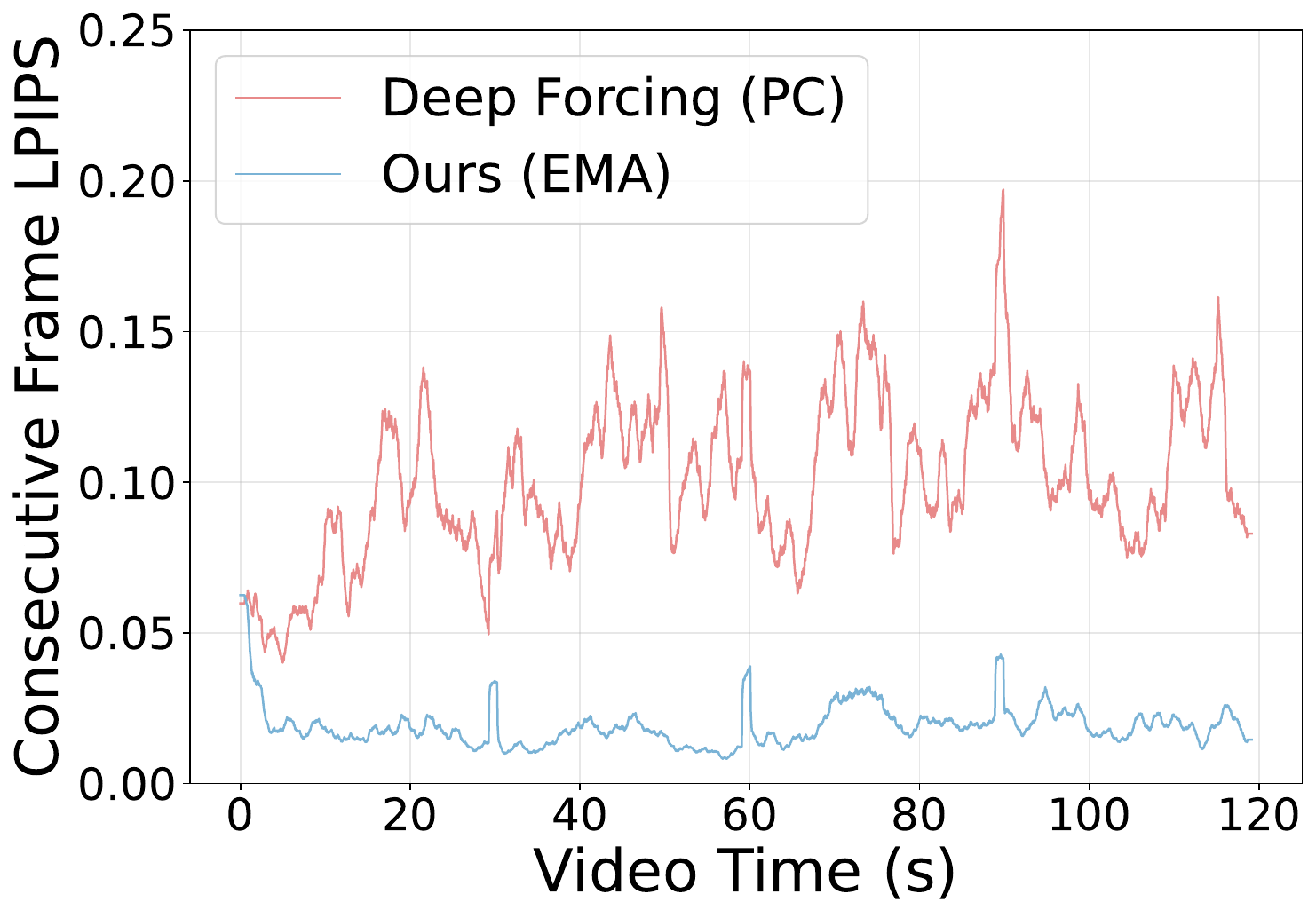}
    \caption{Consecutive LPIPS}
    \label{fig:pc_lpips}
\end{subfigure}
\caption{\textbf{Failure mode of Participative Compression.} \textbf{(a)} Participative Compression (PC), proposed in Deep Forcing~\cite{deepforcing}, rapidly converges to retaining the same long-persisted tokens in the compressed frames, discarding most newly arriving tokens. \textbf{(b)} The few newly admitted tokens carry high attention scores, so each rare cache update exerts a disproportionately strong influence on generation. 
\textbf{(c, d)} This causes frame-to-frame instability: consecutive SSIM drops and LPIPS spikes indicate abrupt visual shifts whenever the cache content changes. Our EMA memory evolves continuously, maintaining smooth transitions.}
\label{fig:pc_failure}
\end{figure*}

% Introducing Memory in video generation (need add more)
To retain past context during extended generation, several approaches adapt sliding windows by leveraging the attention sink phenomenon~\cite{streamingllm}, preserving initial frames as fixed anchors~\cite{rollingforcing,longlive,infinityrope,deepforcing}. 
Deep Forcing~\cite{deepforcing} goes further with Participative Compression, selecting cached tokens by cumulative attention score (\cref{fig:method_comparison}(b)).
However, as shown in \cref{fig:pc_failure}, the selected set rapidly converges to long-persisted tokens, and newly admitted tokens carry disproportionately high scores, causing abrupt visual shifts at each cache update.
% While Deep Forcing~\cite{deepforcing} attempts to improve this by selecting cached tokens dynamically using cumulative attention scores (\cref{fig:method_comparison}(b)), it fails in updating visual shifts for long videos (\cref{fig:pc_failure}). 
% Alternatively, taking inspiration from human dynamic memory, several methods attempt to compress long history videos into short contexts as memory~\cite{yu2025videossm, zhang2025packing, wu2026infinite}; however, these approaches inevitably encounter out-of-memory (OOM) errors when scaling to hours-long generation, and they introduce significant overhead from a context compression module alongside fine-tuning of the base model. 

Current approaches either evict past context entirely or converge to a stagnant token set that shifts abruptly whenever the cache updates.
\emph{None maintains a smoothly evolving representation that adapts as the video unfolds.}

We introduce \textbf{MemRoPE}, a training-free infinite long video generation framework with two co-designed components.
\textbf{Memory Tokens} continuously compress all past keys into dual long-term and short-term streams via exponential moving averages (EMA), maintaining both persistent identity and recent dynamics within a fixed-size cache.
% (\cref{fig:method_comparison}).
This aggregation requires keys free of positional encoding, since merging keys from different timesteps would otherwise mix incompatible rotary phases.
\textbf{Online RoPE Indexing} enables this by storing keys without positional embedding (\cref{fig:method_comparison}(c)) and applying block-relative indices at attention time, which also resolves positional extrapolation. 
MemRoPE is entirely \emph{training-free} and supports \emph{unbounded generation} with constant memory; \cref{fig:one_hour} demonstrates a continuous one-hour example that preserves temporal consistency throughout. Our contributions are as follows:
\begin{itemize}
    \item We propose \textbf{MemRoPE}, a training-free framework for infinite-length video generation that jointly addresses context retention and positional extrapolation, enabling the generation of hours-long videos.
    \item We introduce \textbf{Memory Tokens}, an EMA-based dual memory mechanism that continuously compresses generation history into the long-term and short-term representations within a fixed-size cache.
    \item We present \textbf{Online RoPE Indexing}, which decouples content from position in the KV cache, simultaneously enabling clean temporal aggregation and resolving positional extrapolation without post-hoc corrections.
    \item We demonstrate that MemRoPE outperforms state-of-the-art training-free methods in visual fidelity and consistency across video generation tasks ranging from minute-scale to hour-long sequences.
\end{itemize}

\section{Related Work}
\label{sec:related}

\subsection{Autoregressive Video Generation}
\label{sec:related_arvideo}

Autoregressive video generation spans several paradigms: token-based methods~\cite{videogpt,magvit,magvitv2,videopoet,nova} quantize video for next-token prediction; chunk-level diffusion~\cite{pyramidalflow,magi1,skyreels} denoises multi-frame chunks; and FramePack~\cite{zhang2025packing} and StreamDiT~\cite{streamdit} reduce memory via compressed contexts and window attention.

Most relevant to our work is frame-level autoregressive diffusion.
CausVid~\cite{causvid} distills a bidirectional DiT into a causal generator via DMD, and Self Forcing~\cite{selfforcing} closes the train-test gap by conditioning on self-generated frames.
Self-Forcing++~\cite{selfforcingpp} extends this to four minutes with teacher-guided error correction, Rolling Forcing~\cite{rollingforcing} introduces a rolling denoising window, and Causal Forcing~\cite{causalforcing} improves distillation via ODE initialization.
LongLive~\cite{longlive} aligns training with inference-time rollout for 240-second generation.
FAR~\cite{gu2025long} compresses distant frames via aggressive patchification and proposes FlexRoPE for 16$\times$ temporal extrapolation; while conceptually related to our dual-rate memory, FAR requires training from scratch, whereas MemRoPE is training-free.

\subsection{Long-Horizon Context in Long Video Generation}
\label{sec:related_longhorizon}

Extending autoregressive generation beyond the training horizon raises two challenges: retaining useful past context and maintaining valid positional encoding.
\vspace{-3mm}

\paragraph{Context retention and memory mechanism.}
% KV cache compression has been widely studied for LLMs through attention-based token selection~\cite{zhang2023h2o,li2024snapkv}, token merging~\cite{wang2024model,zhang2024cam}, dynamic budget allocation~\cite{wan2024d2o}, and correlation-aware eviction~\cite{ghadia2025dialogue}, but these techniques operate within a fixed-length cache and cannot recover information once evicted~\cite{zhang2025packing}.
KV cache compression has been widely studied for LLMs through attention-based token selection~\cite{zhang2023h2o,li2024snapkv}, token merging~\cite{wang2024model,zhang2024cam}, dynamic budget allocation~\cite{wan2024d2o}, and correlation-aware eviction~\cite{ghadia2025dialogue}, but these techniques operate within a fixed-length cache and cannot recover information once evicted~\cite{zhang2025packing}.
EMA-based temporal aggregation has also been explored as an architectural primitive~\cite{ma2022mega,ma2024megalodon}, but requires modifications to the model architecture and training from scratch. Our memory tokens instead operate on the KV cache of an unmodified pretrained model.

To retain context beyond the window, StreamingLLM~\cite{streamingllm} identified the \emph{attention sink} phenomenon, and video generation methods adopted a similar strategy of preserving initial frames as static anchors~\cite{rollingforcing, longlive}. Deep Forcing~\cite{deepforcing} goes beyond this with Participative Compression, but the selected tokens quickly collapse to a fixed set during the generation process (\cref{fig:pc_failure}). Other approaches introduce long-context streaming tuning~\cite{contextforcing}, memory modules within the diffusion-forcing paradigm~\cite{streamingt2v, diffusionforcing}, compressed video histories~\cite{zhang2025pretraining, yu2025videossm, zhang2025packing, wu2026infinite}, or geometric priors from 3D representations~\cite{xiao2025worldmem, yu2025context, li2025vmem}. However, these methods either incur growing memory that becomes impractical at hour-length scales or rely on rigid priors that transfer poorly to open-ended synthesis.

In contrast, MemRoPE maintains a fixed-size, continuously evolving memory within a causal autoregressive KV cache, adapting to new content without increasing memory cost regardless of the total generated video length.

\vspace{-2mm}
\paragraph{Positional extrapolation.}
Extending RoPE~\cite{rope} beyond its trained range has been studied in LLMs through position interpolation~\cite{pi}, frequency scaling~\cite{ntk}, and their combinations~\cite{yarn}, but LoL~\cite{lol} shows these induce a quality-dynamics tradeoff in video diffusion.
In the video domain, $\infty$-RoPE~\cite{infinityrope} confines indices to the trained range via a block-relative coordinate system, and RIFLEx~\cite{riflex} targets bidirectional settings.
However, these methods store keys with RoPE already applied and re-rotate them upon eviction; since RoPE does not distribute over addition (\cref{eq:rope_mixing}), averaging such keys from different timesteps produces invalid representations, preventing temporal aggregation (\cref{sec:online_rope}).
Our Online RoPE Indexing instead stores keys without positional encoding and applies block-relative RoPE for the first time at each attention step, which makes EMA-based aggregation well-defined while resolving positional extrapolation.

\begin{figure*}[t]
    \centering
    \includegraphics[width=0.9\textwidth]{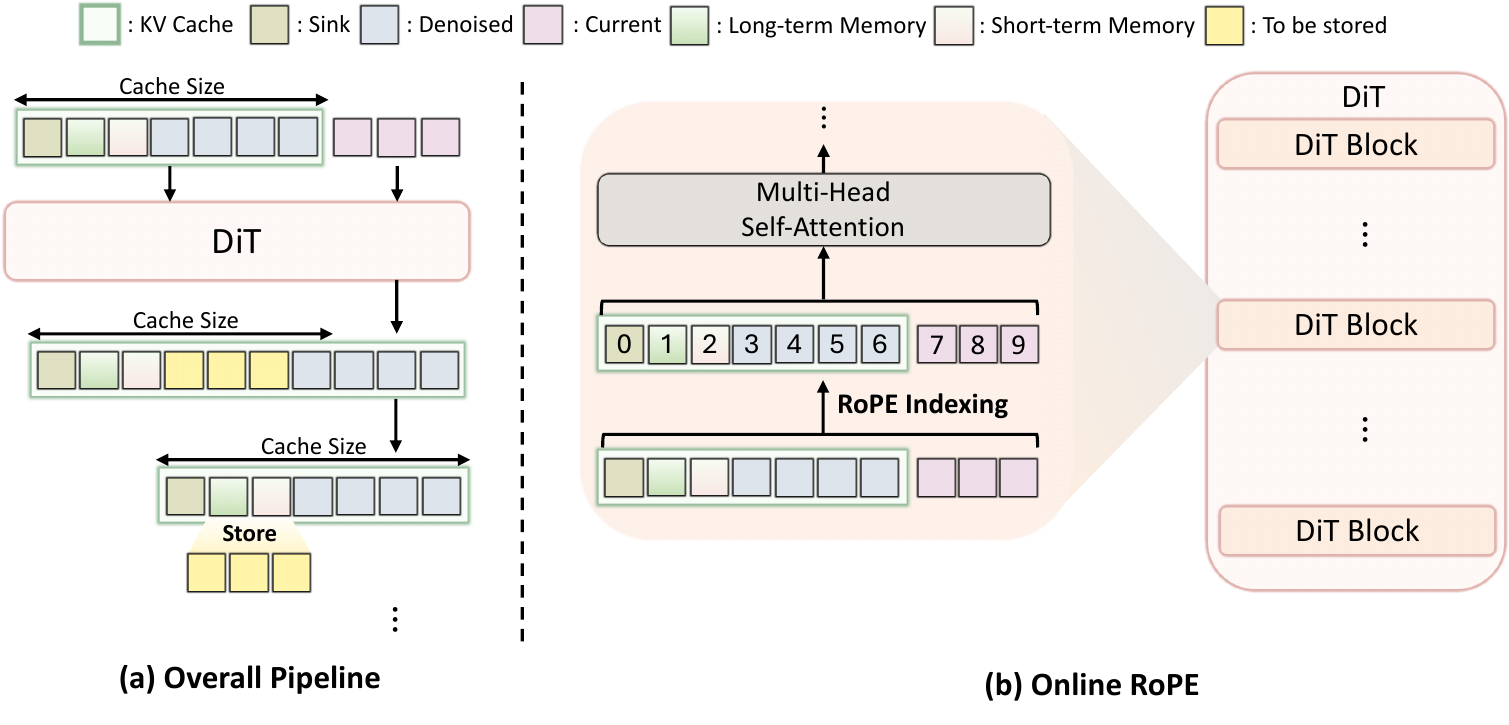}
    \caption{\textbf{Overview of MemRoPE.} \textbf{(a)} At each autoregressive step, the three-tier KV cache (sink, memory, and recent tokens) is concatenated with the current noisy chunk and fed into the DiT. When the local window is full, the oldest frames are absorbed into the long- and short-term memory tokens via dual EMA updates. \textbf{(b)} All cached keys are stored without RoPE, enabling temporal aggregation for memory tokens. At attention time, contiguous block-relative indices are assigned to the full sequence, and RoPE is applied on the fly, ensuring that indices never exceed the training range.}
    \label{fig:method_detail}
\end{figure*}

\section{Background}
\label{sec:prelim}

\paragraph{Autoregressive inference.}
Autoregressive video diffusion generates video one frame chunk at a time~\cite{causvid,selfforcing,longlive,selfforcingpp,lol,rollingforcing,deepforcing}.
Given the KV cache of all previously generated frames $\{x_1, \dots, x_{t-1}\}$, the transformer $\mathcal{G}_\theta$ denoises a noisy input $x_t^{(\sigma)}$ through a small number of diffusion steps:
\begin{equation}
    \hat{x}_t = \mathcal{G}_\theta\!\bigl(x_t^{(\sigma)},\; \sigma,\; \mathbf{K}_{<t},\, \mathbf{V}_{<t}\bigr).
\end{equation}
The resulting key-value pairs $(W_K \hat{x}_t,\, W_V \hat{x}_t)$ are then appended to the KV cache as conditioning for the next step. Because the model uses causal attention, keys and values of past frames can be cached and reused in subsequent generation steps, avoiding redundant computation. Since storing all past KV states is infeasible for long video generations, a sliding window retains only the most recent frames, evicting older ones via FIFO (\cref{fig:method_comparison}(a))~\cite{causvid,selfforcing,longlive,selfforcingpp,lol,rollingforcing,deepforcing}.

\paragraph{Rotary Position Embedding.}
Positional ordering is often encoded via 3D-RoPE in video generation~\cite{rope, wan}, which incorporates the temporal and spatial coordinates of query and key tokens using the following relation:
\begin{equation}
    \tilde{q}_i = R_i\, q_i, \qquad \tilde{k}_j = R_j\, k_j,
\end{equation}
where $R_i \in \mathbb{R}^{d \times d}$ is a rotation matrix encoding the spatiotemporal coordinates of token $i$.
The key property is that the dot product $\tilde{q}_i^\top \tilde{k}_j = q_i^\top R_{i-j}\, k_j$ depends only on the \emph{relative} distance $i - j$, enabling the model to learn position-invariant attention patterns.
In standard practice, RoPE is applied at cache time: each key is stored as $\tilde{k}_j = R_j\, k_j$, permanently binding it to its exact spatiotemporal position $j$. Value states are not rotated and are cached as-is.

\section{Method}
\label{sec:method}

We propose \textbf{MemRoPE}, a training-free framework that is implemented through two co-designed mechanisms: \textbf{Memory Tokens} (\cref{sec:memory}) and \textbf{Online RoPE Indexing} (\cref{sec:online_rope}). \cref{fig:method_detail} provides a comprehensive visual overview.

\subsection{Memory Tokens}
\label{sec:memory}

\paragraph{Why existing cache management fails.}
FIFO eviction irrecoverably discards past frames, causing identity drift, scene inconsistency, and motion stagnation over long horizons~\cite{selfforcing,longlive,deepforcing,infinityrope}.
To mitigate this, several methods~\cite{rollingforcing,longlive,deepforcing} retain the initial frames as \emph{attention sinks} alongside the sliding window, inspired by the attention sink phenomenon in LLMs~\cite{streamingllm}.
This is well motivated, as early frames lie within the training horizon and tend to be the highest quality in DMD-distilled models; however, these sinks are frozen at generation start and cannot reflect later changes in scene content or subject identity.

Deep Forcing~\cite{deepforcing} extends this with \emph{Participative Compression} (PC), which dynamically selects cached tokens by cumulative attention score.
However, the selection mechanism suffers from a self-reinforcing bias: tokens that have persisted in the cache accumulate higher attention scores, making them increasingly likely to be retained regardless of their current relevance.
As shown in \cref{fig:pc_failure}(a), the selected set rapidly converges to a stagnant, fixed group of long-persisted tokens. Moreover, when a new token is finally admitted, it carries a disproportionately high attention score relative to the entrenched set (\cref{fig:pc_failure}(b)), producing abrupt, highly noticeable visual shifts at each cache update (\cref{fig:pc_failure}(c,d)).

Across all existing methods, preserved context is either a static snapshot of early frames or subject to discrete token selection that converges to a stale set.
We instead propose a \emph{continuously evolving} representation.

\vspace{-7mm}
\paragraph{Dual EMA memory.}
We propose Memory Tokens: a fixed-size buffer evolving throughout generation via exponential moving averages (EMA). As chunks exit the window, their spatially pooled key $\bar{k}_{\mathrm{new}}$ updates two parallel streams:
\begin{align}
    \mu_L^{(t)} &= (1 - \alpha_L)\, \mu_L^{(t-1)} + \alpha_L\, \bar{k}_{\mathrm{new}}, \label{eq:ema_long} \\
    \mu_S^{(t)} &= (1 - \alpha_S)\, \mu_S^{(t-1)} + \alpha_S\, \bar{k}_{\mathrm{new}}, \label{eq:ema_short}
\end{align}
% where $\alpha_L = 0.01$ (slow decay) accumulates the full generation history into a stable representation that consistently attends to the primary subject across the video, and $\alpha_S = 0.1$ (fast decay) tracks recent frames, responding more to temporally varying regions such as background changes and camera motion.
where $\alpha_L \ll \alpha_S$: the long-term stream accumulates the full generation history into a stable representation, while the short-term stream tracks recent dynamics.
The same update applies to value states.
Memory size is {constant} regardless of video length, enabling unbounded generation within a fixed budget.

\subsection{Online RoPE Indexing}
\label{sec:online_rope}

% \paragraph{The aggregation barrier.}

The memory token updates in \cref{eq:ema_long,eq:ema_short} average keys from different timesteps.
However, conventional KV caches store keys with RoPE already applied: $\tilde{k}_j = R_j k_j$.
Averaging such keys mixes incompatible rotary phases:
\begin{equation}
    \alpha\, R_j k_j + (1{-}\alpha)\, R_{j'} k_{j'} \;\neq\; R_\phi\!\bigl(\alpha\, k_j + (1{-}\alpha)\, k_{j'}\bigr)
    \label{eq:rope_mixing}
\end{equation}
for any rotation $R_\phi$, since RoPE does not distribute over addition.
As a consequence, the averaged key no longer corresponds to any valid temporal position, breaking the relative-position structure that attention relies on.

We resolve this aggregation barrier together with the positional extrapolation problem through a single design change: \emph{store all keys without RoPE, and dynamically assign block-relative temporal indices at each attention step.}

\vspace{-2mm}
\paragraph{Position-free caching.}
Every individual key is computed in its raw, unrotated form without positional embeddings before entering the cache:
\begin{equation}
    k_j^{\mathrm{cache}} = k_j \qquad (\text{no RoPE}).
    \label{eq:unroped}
\end{equation}
EMA updates (\cref{eq:ema_long,eq:ema_short}) aggregate these position-free keys in the cache, and the resulting memory tokens operate as valid key vectors that can dynamically receive any positional index at attention time.

\vspace{-2mm}

\paragraph{Block-relative index assignment.}
At each generation step, we treat the cache as a single block and assign a contiguous index map $\phi$ starting from zero:
\begin{equation}
\resizebox{0.9\columnwidth}{!}{$
    \underbrace{[0,\, \dots,\, S{-}1]}_{\mathrm{sink}} \;\;
    \underbrace{[S,\, \dots,\, S{+}2M{-}1]}_{\mathrm{memory}} \;\;
    \underbrace{[S{+}2M,\, \dots,\, S{+}2M{+}L{-}1]}_{\mathrm{local}}
$}
\label{eq:index}
\end{equation}
where $S$, $M$, $L$ are the sink, memory, and local window sizes.
The current query receives indices starting from $S{+}2M{+}L$.
RoPE is then applied on the fly: $\tilde{k}_j = R_{\phi(j)}\, k_j^{\mathrm{cache}}$ for every cached key, and $\tilde{q}_i = R_{\phi(i)}\, q_i$ for every current query.
Value states are not rotated.
Since all indices are recomputed from zero at every generation step, no position ever exceeds the total cache size $C = S{+}2M{+}L$, which remains within the range seen during training.

\vspace{-3mm}

\paragraph{Relationship to block-relative RoPE.}
$\infty$-RoPE~\cite{infinityrope} re-anchors cached keys backward at each step to keep the newest block at the maximum trained index, requiring per-step phase rotations on keys already stored with RoPE.
Our Online RoPE Indexing instead stores keys without RoPE and applies positional encoding once at attention time, eliminating per-step rotations and enabling temporal aggregation via EMA, which is ill-defined over keys stored with conflicting rotary phases (\cref{eq:rope_mixing}).
We map the cache compactly to $[0,\, C{-}1]$ rather than $[f_{\mathrm{limit}} - C,\, f_{\mathrm{limit}}]$, where $f_{\mathrm{limit}}$ is the maximum position index seen during pre-training, so that the sink tokens always receive the lowest indices and memory tokens occupy consistent positions in the sequence.
% We map the cache compactly to $[0,\, C{-}1]$ rather than $[f_{\mathrm{limit}} - C,\, f_{\mathrm{limit}}]$.
% We map the cache compactly to $[0,\, C{-}1]$ rather than $[f_{\mathrm{limit}} - C,\, f_{\mathrm{limit}}]$, where $f_{\mathrm{limit}}$ is the maximum position index seen during pre-training.

\begin{algorithm}[t]
\caption{MemRoPE Inference}
\label{alg:memrope}
\begin{algorithmic}[1]
\small
\Require Diffusion Transformer $\mathcal{G}_\theta$, denoising steps $N$, EMA rates $\alpha_L, \alpha_S$
\Statex \textbf{Notation:} $x_t^{(s)}$: noisy latent of chunk $t$ at denoising step $s$; $\hat{x}_t = x_t^{(N)}$: final denoised chunk
\State $\hat{x}_0 \gets \mathcal{G}_\theta(x_0^{(0)}, N)$
\State $\mathbf{K}_{\mathrm{sink}}, \mathbf{V}_{\mathrm{sink}} \gets W_K \hat{x}_0, W_V \hat{x}_0$ \Comment{w/o RoPE}
\State $\mu_L^k, \mu_L^v, \mu_S^k, \mu_S^v \gets \mathbf{0}$; $\mathbf{K}_{\mathrm{local}}, \mathbf{V}_{\mathrm{local}} \gets \emptyset$
\For{$t = 1, 2, \dots$} \Comment{chunk loop}
    \For{$s = 1, \dots, N$} \Comment{denoising steps}
        \State $q_s, k_s, v_s \gets W_Q x_t^{(s)}, W_K x_t^{(s)}, W_V x_t^{(s)}$
        \State $\mathbf{K} \gets [\mathbf{K}_{\mathrm{sink}} \| \mu_L^k \| \mu_S^k \| \mathbf{K}_{\mathrm{local}} \| k_s]$ \Comment{w/o RoPE}
        \State $\mathbf{V} \gets [\mathbf{V}_{\mathrm{sink}} \| \mu_L^v \| \mu_S^v \| \mathbf{V}_{\mathrm{local}} \| v_s]$
        \State $\phi \gets [0, \dots, |\mathbf{K}|{-}1]$ \Comment{Online RoPE}
        \State Apply $R_\phi$ to $q_s, \mathbf{K}$
        \State $x_t^{(s+1)} \gets \mathrm{DiT\text{-}Block}(R_{\phi} q_s, R_{\phi} \mathbf{K}, \mathbf{V})$
    \EndFor
    \State $\hat{x}_t \gets x_t^{(N)}$ \Comment{denoised chunk}
    \State Append $W_K \hat{x}_t, W_V \hat{x}_t$ to $\mathbf{K}_{\mathrm{local}}, \mathbf{V}_{\mathrm{local}}$ \Comment{w/o RoPE}
    \If{$|\mathbf{K}_{\mathrm{local}}| > L$} \Comment{evict \& update memory}
        \State $\bar{k} \gets \mathrm{SpatialPool}(\mathbf{K}_{\mathrm{local}}[0])$
        \State $\bar{v} \gets \mathrm{SpatialPool}(\mathbf{V}_{\mathrm{local}}[0])$
        \State $\mu_L^k \gets (1{-}\alpha_L)\mu_L^k + \alpha_L\bar{k}$
        \State $\mu_L^v \gets (1{-}\alpha_L)\mu_L^v + \alpha_L\bar{v}$
        \State $\mu_S^k \gets (1{-}\alpha_S)\mu_S^k + \alpha_S\bar{k}$
        \State $\mu_S^v \gets (1{-}\alpha_S)\mu_S^v + \alpha_S\bar{v}$
        \State Evict oldest chunk from $\mathbf{K}_{\mathrm{local}}, \mathbf{V}_{\mathrm{local}}$
    \EndIf
\EndFor
\end{algorithmic}
\end{algorithm}

\definecolor{firstbg}{HTML}{A5D6A7}  % dark green background for 1st
\definecolor{secondbg}{HTML}{E8F5E9} % light green background for 2nd
\definecolor{groupbg}{HTML}{F5F5F5}   % very very light gray for group header

% Commands for colored cells
\newcommand{\first}[1]{\cellcolor{firstbg}\textbf{#1}}
\newcommand{\second}[1]{\cellcolor{secondbg}{#1}}

% Required packages (add to preamble):
% \usepackage{colortbl, xcolor, booktabs, makecell, array}

\begin{table*}[t]
\centering
\caption{\textbf{Quantitative comparison on long video generation.} We report VBench-Long~\cite{vbench} metrics. Within each base model group, best results are {bold with darker background} and second best are with lighter background. $\Delta$ denotes the average improvement over the respective base model.}
\label{tab:main}
\resizebox{\textwidth}{!}{%
\begin{tabular}{l cccccc !{\vrule width 0.8pt} c c}
\toprule
Method & \makecell{Aesthetic\\Quality} $\uparrow$ & \makecell{Background\\Consistency} $\uparrow$ & \makecell{Imaging\\Quality} $\uparrow$ & \makecell{Motion\\Smoothness} $\uparrow$ & \makecell{Subject\\Consistency} $\uparrow$ & \makecell{Temporal\\Flickering} $\uparrow$ & Average $\uparrow$ & $\Delta$ \\
\midrule
\multicolumn{9}{c}{\textit{120 seconds}} \\
\midrule
\rowcolor{groupbg}
\multicolumn{9}{l}{\textit{Results on Self-Forcing}~\cite{selfforcing}} \\
Self-Forcing~\cite{selfforcing}            & 55.11 & 95.41 & \second{65.95} & 97.17 & 95.35 & \second{96.53} & \second{84.25} & -- \\
Deep Forcing~\cite{deepforcing}            & \first{56.86} & 94.58 & 65.02 & 97.07 & 94.17 & 95.33 & 83.84 & $-$0.41 \\
$\infty$-RoPE~\cite{infinityrope}          & 52.82 & \second{95.66} & 61.07 & \first{98.48} & \second{96.30} & \first{97.50} & 83.64 & $-$0.61 \\
\textbf{MemRoPE (Ours)}                           & \second{56.77} & \first{95.54} & \first{68.44} & \second{97.90} & \first{96.37} & {96.33} & \first{85.23} & \first{$+$0.98} \\
\addlinespace[2pt]
\rowcolor{groupbg}
\multicolumn{9}{l}{\textit{Results on LongLive}~\cite{longlive}} \\
LongLive~\cite{longlive}                   & 57.21 & 95.96 & 67.13 & 98.56 & 97.06 & 97.12 & 85.51 & -- \\
Deep Forcing~\cite{deepforcing}            & \second{59.16} & 96.13 & \second{68.07} & 98.51 & 96.79 & 97.38 & \second{86.01} & \second{$+$0.50} \\
$\infty$-RoPE~\cite{infinityrope}          & 57.87 & \first{96.46} & 64.84 & \first{99.00} & \second{97.29} & \first{98.00} & 85.58 & $+$0.07 \\
\textbf{MemRoPE (Ours)}                           & \first{59.25} & \second{96.29} & \first{69.43} & \second{98.73} & \first{97.54} & \second{97.47} & \first{86.45} & \first{$+$0.94} \\
\midrule
\multicolumn{9}{c}{\textit{240 seconds}} \\
\midrule
\rowcolor{groupbg}
\multicolumn{9}{l}{\textit{Results on Self-Forcing}~\cite{selfforcing}} \\
Self-Forcing~\cite{selfforcing}            & 51.00 & 95.01 & \second{61.52} & \second{98.18} & 95.83 & \second{96.72} & \second{83.04} & -- \\
Deep Forcing~\cite{deepforcing}            & \second{52.20} & 93.91 & 59.50 & 96.72 & 92.28 & 95.38 & 81.66 & $-$1.38 \\
$\infty$-RoPE~\cite{infinityrope}          & 50.51 & \first{95.52} & {58.81} & \first{98.47} & \second{96.24} & \first{97.48} & {82.84} & $-$0.20 \\
\textbf{MemRoPE (Ours)}                           & \first{55.54} & \second{95.45} & \first{67.77} & {97.93} & \first{96.30} & {96.39} & \first{84.89} & \first{+1.85} \\
\addlinespace[2pt]
\rowcolor{groupbg}
\multicolumn{9}{l}{\textit{Results on LongLive}~\cite{longlive}} \\
LongLive~\cite{longlive}                   & 56.70 & 95.80 & 66.90 & 98.52 & 97.02 & 97.04 & 85.33 & -- \\
Deep Forcing~\cite{deepforcing}            & \second{57.75} & 96.03 & \second{66.48} & 98.50 & 96.65 & \second{97.47} & \second{85.48} & \second{$+$0.15} \\
$\infty$-RoPE~\cite{infinityrope}          & 57.47 & \second{96.34} & 63.38 & \first{98.99} & \second{97.11} & \first{97.94} & 85.21 & $-$0.12 \\
\textbf{MemRoPE (Ours)}                           & \first{58.90} & \first{96.39} & \first{68.93} & \second{98.59} & \first{97.37} & {97.13} & \first{86.22} & \first{$+$0.89} \\
\bottomrule
\end{tabular}%
}
\end{table*}

\vspace{-4mm}

\paragraph{Relationship to Dynamic RoPE.}
Rolling Forcing~\cite{rollingforcing} stores raw keys for its sink tokens and reapplies RoPE dynamically at attention time, an approach conceptually close to our Online RoPE Indexing. However, this is limited to the static sink frames; keys in the rolling window are still stored with monotonically increasing RoPE indices, which both precludes temporal aggregation and reintroduces positional extrapolation as the generated sequence grows. We generalize position-free storage to the entire cache, including the sink, memory, and local window alike. This makes EMA aggregation well-defined across all cache slots while keeping all indices within the training range.

\subsection{Three-Tier Cache}
\label{sec:cache_arch}

The complete MemRoPE cache at step $t$ is:
\begin{equation}
    \mathbf{K}^{(t)} = \bigl[\;\underbrace{\mathbf{K}_{\mathrm{sink}}}_{S} \;\|\; \underbrace{\mathbf{K}_{\mathrm{mem}}}_{2M} \;\|\; \underbrace{\mathbf{K}_{\mathrm{local}}}_{L}\;\bigr],
    \label{eq:cache}
\end{equation}
with the value cache $\mathbf{V}^{(t)}$ structured identically.
This ordering mirrors the temporal structure of the video: $\mathbf{K}_{\mathrm{sink}}$ anchors the earliest high-quality frames, $\mathbf{K}_{\mathrm{mem}} = [\mu_L \| \mu_S]$ summarizes the evolving history via dual EMA, and $\mathbf{K}_{\mathrm{local}}$ captures the recent sliding window.
All keys are stored position-free (\cref{fig:method_comparison}); block-relative RoPE indices (\cref{eq:index}) are applied at every attention call.

The full inference procedure is given in \cref{alg:memrope}.

\begin{figure*}[t]
    \centering
    \includegraphics[width=\textwidth]{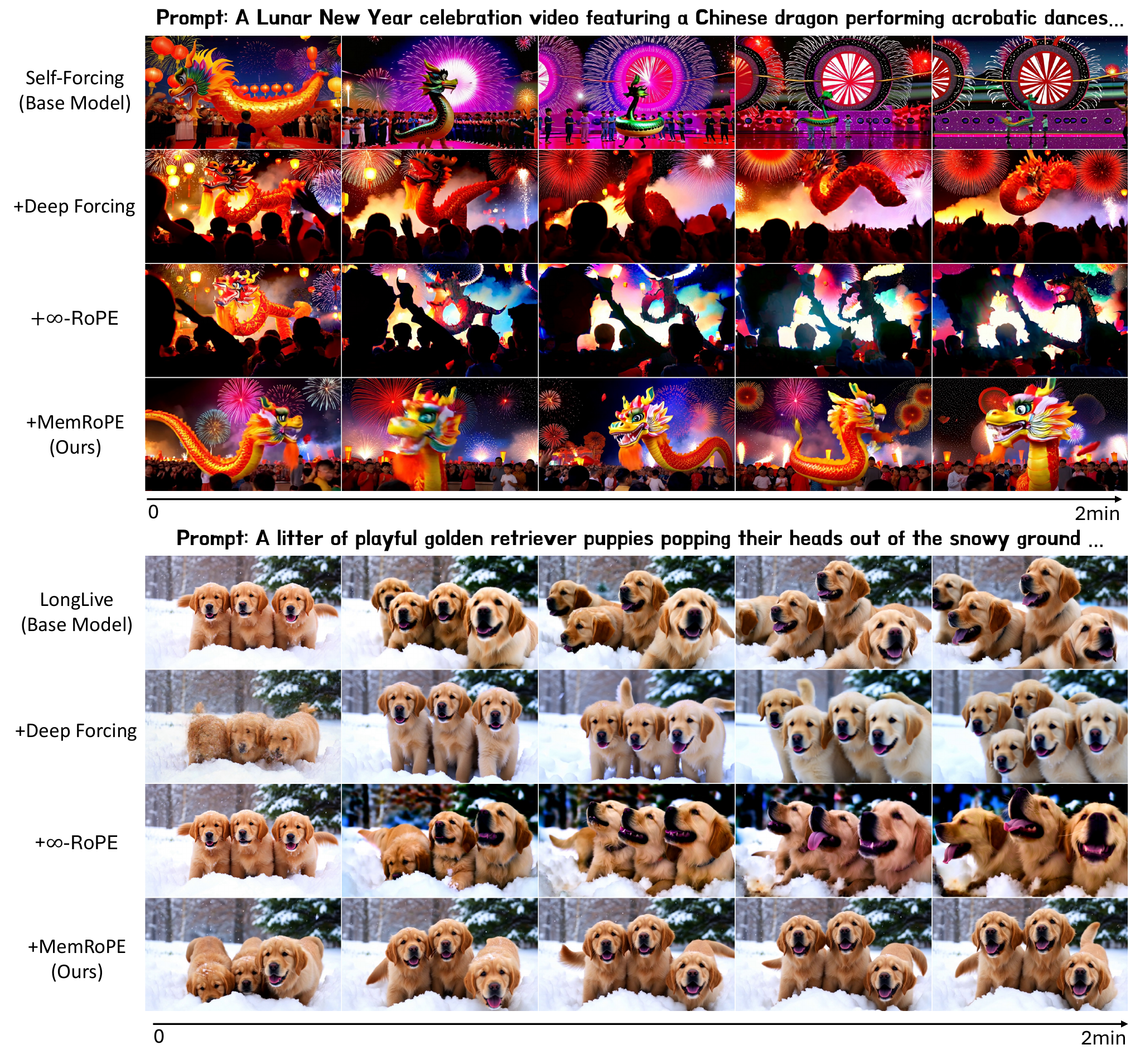}
    \caption{Qualitative comparison on 2-minute video generation. MemRoPE maintains subject identity and background consistency throughout, whereas baselines exhibit progressive degradation, including structural collapse and color corruption.}
    \label{fig:qualitative}
\end{figure*}

\section{Experiments}
\label{sec:exp}

\subsection{Setup}
\label{sec:setup}

\paragraph{Implementation details.}
We build on the Wan2.1-T2V-1.3B architecture~\cite{wan} and evaluate MemRoPE on two base models: Self-Forcing~\cite{selfforcing} and LongLive~\cite{longlive}.
% , which applies streaming long tuning on top of Self-Forcing to close the train-test gap for long-horizon generation.
Since MemRoPE is training-free, it can be plugged into any autoregressive video diffusion model without modification.
Video is generated autoregressively in chunks of 3 latent frames with a 4-step denoising schedule at timesteps $\{1000, 750, 500, 250\}$.
The three-tier cache consists of $S=3$ sink tokens, $M=1$ memory token per stream (long-term and short-term), and a local window of $L=4$ frames, with EMA decay rates $\alpha_L = 0.01$ and $\alpha_S = 0.1$.
The KV cache is updated only after the final denoising step of each chunk, ensuring that only fully denoised keys and values enter the cache.
Deep Forcing~\cite{deepforcing} and $\infty$-RoPE~\cite{infinityrope} are also training-free methods; we apply them on both base models under the same protocol for fair comparison.

% 
% \noindent\textbf{Baselines.}
% All baselines operate on the same Wan2.1-T2V-1.3B backbone under the autoregressive diffusion framework.
% We compare against Self-Forcing~\cite{selfforcing}, Rolling Forcing~\cite{rollingforcing}, Deep Forcing~\cite{deepforcing}, LongLive~\cite{longlive}, and $\infty$-RoPE~\cite{infinityrope}.

\vspace{-5mm}
\paragraph{Evaluation protocol.}
We generate videos from text prompts sampled from MovieGenBench~\cite{moviegen} at $480 \times 832$ resolution and 16 fps.
For 120 and 240 seconds, we use 128 prompts; for 480 seconds, 20 randomly sampled prompts; for 1 hour, 10 randomly sampled prompts; and for ablation studies, 20 randomly sampled prompts at 60 seconds.
Following Deep Forcing~\cite{deepforcing} and Self-Forcing~\cite{selfforcing}, all prompts are refined using Qwen2.5-7B-Instruct.
% Following prior work~\cite{deepforcing,selfforcing}, all prompts are refined using Qwen2.5-7B-Instruct before generation.
We report metrics from VBench-Long~\cite{vbench}, and further validate with a user study following the 2AFC protocol~\cite{deepforcing} and VLM-based evaluation using Gemini 3.1-Pro~\cite{comanici2025gemini}.

\begin{table*}[t]
\centering
\caption{\textbf{Ultra-long video generation.} We report VBench-Long~\cite{vbench} metrics. Within each base model group, best results are {bold with darker background}.}
\label{tab:ultra_long}
\resizebox{0.9\textwidth}{!}{%
\begin{tabular}{l cccccc !{\vrule width 0.8pt} c}
\toprule
Method & \makecell{Aesthetic\\Quality} $\uparrow$ & \makecell{Background\\Consistency} $\uparrow$ & \makecell{Imaging\\Quality} $\uparrow$ & \makecell{Motion\\Smoothness} $\uparrow$ & \makecell{Subject\\Consistency} $\uparrow$ & \makecell{Temporal\\Flickering} $\uparrow$ & Average $\uparrow$ \\
\midrule
\multicolumn{8}{c}{\textit{480 seconds}} \\
\midrule
\rowcolor{groupbg}
\multicolumn{8}{l}{\textit{Results on Self-Forcing}~\cite{selfforcing}} \\
$\infty$-RoPE~\cite{infinityrope}          & 48.35 & {95.45} & 56.00 & \first{98.59} & {96.46} & \first{97.71} & 82.09 \\
\textbf{MemRoPE (Ours)}                           & \first{56.12} & \first{95.65} & \first{68.23} & 97.96 & \first{96.87} & 96.42 & \first{85.21} \\
\addlinespace[2pt]
\rowcolor{groupbg}
\multicolumn{8}{l}{\textit{Results on LongLive}~\cite{longlive}} \\
$\infty$-RoPE~\cite{infinityrope}          & 56.77 & \first{96.25} & 61.32 & \first{99.05} & 97.16 & \first{97.95} & 84.75 \\
\textbf{MemRoPE (Ours)}                           & \first{57.96} & 96.12 & \first{67.52} & 98.66 & \first{97.37} & 97.23 & \first{85.81} \\
\midrule
\multicolumn{8}{c}{\textit{1 hour}} \\
\midrule
\rowcolor{groupbg}
\multicolumn{8}{l}{\textit{Results on LongLive}~\cite{longlive}} \\
$\infty$-RoPE~\cite{infinityrope}          & 60.87 & 96.42 & 70.42 & 99.01 & 97.71 & 97.62 & 87.01 \\
\textbf{MemRoPE (Ours)}                           & \first{63.05} & \first{96.77} & \first{72.71} & \first{99.07} & \first{98.18} & \first{98.14} & \first{87.99} \\
\bottomrule
\end{tabular}%
}
\end{table*}

\subsection{Results}
\label{sec:results}

% We evaluate MemRoPE through the following research questions:
% (1) Does MemRoPE improve quantitative metrics over existing methods across different generation durations? (\cref{sec:quantitative})
% (2) Are the improvements visually perceptible in generated videos? (\cref{sec:qualitative})
% (3) Can MemRoPE scale to ultra-long generation beyond the training horizon? (\cref{sec:ultra_long})
% (4) Do human evaluators and VLMs confirm the quality gains? (\cref{sec:human_vlm})
% (5) How do individual memory components and hyperparameters affect performance? (\cref{sec:ablations})

% \paragraph{Quantitative results.}
% \label{sec:quantitative}
% \cref{tab:main} summarizes the VBench-Long scores across three generation durations. MemRoPE achieves the highest average score at every duration, outperforming all baselines in Aesthetic Quality, Imaging Quality, Motion Smoothness, and Subject Consistency. These gains are most pronounced at 240 seconds, where the average gap over the second-best method widens to 2.7 points, indicating that the evolving memory mechanism becomes increasingly beneficial as the video grows longer. Conversely, Deep Forcing records substantially lower temporal consistency and visual quality metrics across all tested durations despite its high Dynamic Degree, suggesting that its elevated motion scores reflect generation instability rather than meaningful dynamics.

\paragraph{Quantitative results.}
\cref{tab:main} summarizes the quantitative comparison at 120 and 240 seconds on two base models (\textit{i.e.}, Self-Forcing, LongLive).
MemRoPE achieves the highest average score across all settings, consistently improving over both base models.
In contrast, Deep Forcing and $\infty$-RoPE degrade the Self-Forcing base at both durations, with Deep Forcing falling 1.38 points below at 240 seconds.
On LongLive, Deep Forcing provides moderate gains at 120 seconds but nearly vanishes at 240 seconds, while $\infty$-RoPE remains near or slightly below the baseline. MemRoPE maintains stable improvement at both durations.
Per-metric analysis reveals that MemRoPE leads in Aesthetic Quality, Imaging Quality, and Subject Consistency---the dimensions most sensitive to long-range context retention---while $\infty$-RoPE tends to score higher on Motion Smoothness and Temporal Flickering, which primarily measure local frame-to-frame stability.
This pattern suggests that $\infty$-RoPE preserves short-range smoothness but lacks the evolving memory needed to maintain visual fidelity and identity over longer horizons.

% \paragraph{Qualitative results.}
% \label{sec:qualitative}
% \cref{fig:qualitative} presents frame-by-frame comparisons over 2-minute generations.
% Deep Forcing~\cite{deepforcing} exhibits early-frame artifacts and progressive color drift; in the first prompt, the snowy background gradually vanishes, and in the second, limb structure breaks down mid-sequence.
% LongLive~\cite{longlive} maintains initial quality but diverges over time---the number of puppies changes and severe blur accumulates in the running sequence, eroding fidelity.
% $\infty$-RoPE~\cite{infinityrope} suffers from the most aggressive error accumulation, with color instability and identity corruption rendering later frames unrecognizable in both prompts.
% MemRoPE preserves subject count, identity, and background consistency throughout, producing the most temporally stable results across both examples.

\vspace{-3mm}
\paragraph{Qualitative results.}
\cref{fig:qualitative} presents frame-by-frame comparisons over 2-minute generations on both base models.
On Self-Forcing, the base model degrades to the point where subjects become unrecognizable by the end of generation. Deep Forcing retains rough shapes but loses all fine detail, while $\infty$-RoPE suffers from severe color instability with no meaningful preservation of the original appearance.
On LongLive, both the base model and Deep Forcing fail to maintain subject consistency---the number of puppies fluctuates from three to four or five across frames, and Deep Forcing additionally shifts the overall color tone. $\infty$-RoPE exhibits the most aggressive error accumulation, with the background collapsing into unrecognizable forms.
MemRoPE preserves subject count, identity, and background consistency throughout on both base models, producing the most temporally stable results across both prompts.

% \paragraph{Ultra-long generation.}
% \label{sec:ultra_long}
% Standard autoregressive video diffusion is bounded by two architectural limits: the 1024-frame latent sequence length imposed by RoPE, and the memory footprint of VAE decoding.
% Among our baselines, only $\infty$-RoPE natively handles positional extrapolation; the remaining methods are capped at 4 minutes and 15 seconds.
% To enable a fair comparison at 480 seconds, we equip all methods with chunked VAE decoding (120 frames per chunk), local attention following the LoL~\cite{lol} framework, and our Online RoPE Indexing to resolve positional extrapolation, so that the only remaining differentiator is each method's KV cache management strategy.
% \cref{tab:ultra_long} reports the results.
% Even at this extreme duration, MemRoPE maintains the highest average score and shows minimal degradation compared to its 240-second results, confirming that the dual EMA memory scales gracefully beyond the original positional limit.

\vspace{-3mm}
\paragraph{Ultra-long generation.}
\cref{tab:ultra_long} extends the comparison to 480 seconds and 1 hour.
Besides context retention, ultra-long video generation is also constrained by the length of RoPE.
The maximum generation length of both Self-Forcing and LongLive is 4 minutes and 15 seconds, limited by the 1024-frame latent sequence.
Among our baselines, only $\infty$-RoPE resolves this through a block-relative coordinate system that confines indices to the trained range.
Online RoPE Indexing similarly confines indices to the trained range, enabling both MemRoPE and $\infty$-RoPE to generate beyond this limit.

% At 480 seconds, MemRoPE outperforms $\infty$-RoPE by over 3 points on Self-Forcing and 1 point on LongLive in average score.
% The gap is driven by Aesthetic Quality and Imaging Quality, where MemRoPE leads by a large margin, while $\infty$-RoPE again scores higher only on Motion Smoothness and Temporal Flickering.
% This is consistent with the pattern observed at shorter durations: $\infty$-RoPE maintains local smoothness but cannot retain visual fidelity without an evolving memory. Furthermore, at 1 hour on LongLive, MemRoPE still leads with an average score of 85.97 versus 84.79, demonstrating that the dual EMA memory scales gracefully to hour-length generation.
At 480 seconds, MemRoPE outperforms $\infty$-RoPE by over 3 points on Self-Forcing and 1 point on LongLive in average score.
Consistent with shorter durations, the gap is driven by Aesthetic Quality and Imaging Quality, while $\infty$-RoPE scores higher on Motion Smoothness and Temporal Flickering on both base models.
At 1 hour on LongLive, MemRoPE leads across all six metrics with an average of 87.99 versus 87.01, demonstrating that the dual EMA memory scales gracefully to hour-length generation.

\begin{table}[h]
\centering
\caption{User study (\% favoring Ours).}
\label{tab:user_study}
\scriptsize
\setlength{\tabcolsep}{2pt}
\resizebox{\columnwidth}{!}{%
\begin{tabular}{l cccccc}
\toprule
Baseline & \makecell{Color\\Cons.} & \makecell{Subject\\Cons.} & \makecell{Bg.\\Cons.} & \makecell{Text\\Align.} & \makecell{Motion\\Smo.} & \makecell{Overall\\Pref.} \\
\midrule
Self-Forcing~\cite{selfforcing}       & 93.0 & 93.0 & 97.4 & 91.6   & 95.7 & 98.3 \\
Rolling Forcing~\cite{rollingforcing} & 90.4 & 77.4 & 79.1 & 80.7   & 88.7 & 82.6 \\
\midrule
\rowcolor{groupbg}
\multicolumn{7}{l}{\textit{Results on LongLive}~\cite{longlive}} \\
LongLive~\cite{longlive}             & 81.6 & 70.2 & 83.3 & 66.1   & 71.9 & 71.1 \\
Deep Forcing~\cite{deepforcing}       & 79.0 & 75.2 & 81.9 & 70.5 & 72.4 & 72.6 \\
$\infty$-RoPE~\cite{infinityrope}     & 81.3 & 76.4 & 77.6 & 74.5 & 73.8 & 70.1 \\
\bottomrule
\end{tabular}
}%
\end{table}

% \input{src/table3}

% To complement the automated metrics, we conducted a user study with 30 independent participants following the standard Two-Alternative Forced Choice (2AFC) protocol~\cite{deepforcing}. All participant data was handled with strict anonymity; identifying information was never shared with any third party and was securely deleted afterward. Each participant was asked to review 20 video trials. For each trial, participants were shown two 120-second videos generated from the same prompt, one from MemRoPE and one from a baseline; then they were asked to select the better video in terms of subject consistency, visual quality, motion smoothness, and overall preference. \cref{tab:user_study} reports the preference rates. Participants consistently preferred MemRoPE across all evaluated aspects, strongly corroborating the previously discussed automated quantitative results and confirming that the quality gains are perceptually meaningful.
\vspace{-3mm}
\paragraph{User study.}
\label{sec:human_vlm}
To complement the automated metrics, we conducted a user study with 30 participants following the 2AFC protocol~\cite{deepforcing}; all data was collected anonymously and handled in accordance with ethical guidelines. Each participant compared 20 pairs of 120-second videos (MemRoPE vs.\ a baseline) across six perceptual dimensions listed in \cref{tab:user_study}. Participants consistently preferred MemRoPE across all aspects, corroborating the quantitative results.

% \begin{table}[h!]
% \centering
% \caption{Visual stability scored by Gemini 3.1-Pro~\cite{gemini31pro} on 120-second videos.}
% \label{tab:vlm_eval}
% \scriptsize
% \resizebox{0.55\columnwidth}{!}{%
% \begin{tabular}{l c}
% \toprule
% Method & \makecell{Visual\\Stability} \\
% \midrule
% Self-Forcing~\cite{selfforcing}       & 1.55 \\
% Rolling Forcing~\cite{rollingforcing} & 3.40 \\
% \midrule
% \rowcolor{groupbg}
% \multicolumn{2}{l}{\textit{Results on LongLive}~\cite{longlive}} \\
% LongLive~\cite{longlive}             & 4.10 \\
% Deep Forcing~\cite{deepforcing}       & 3.90 \\
% $\infty$-RoPE~\cite{infinityrope}     & 4.05 \\
% \textbf{MemRoPE (Ours)}              & \textbf{4.15} \\
% \bottomrule
% \end{tabular}
% }%
% \end{table}

\begin{figure*}[t]
    \centering
    \includegraphics[width=\textwidth]{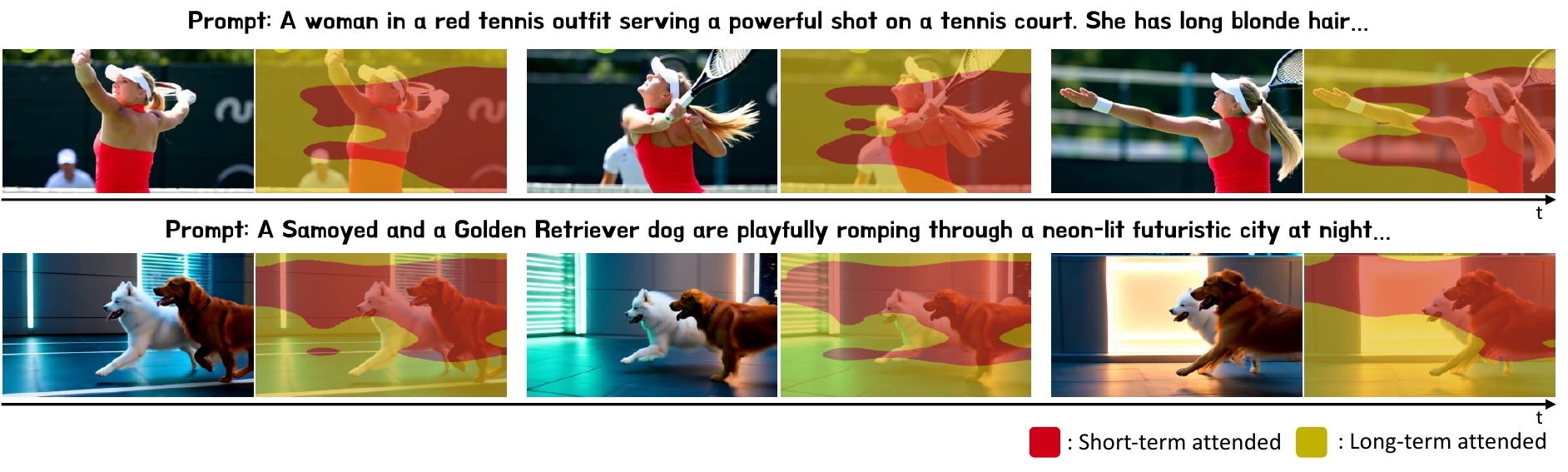}
    \caption{Spatial attention distribution between the dual memory streams. Yellow indicates higher attention to long-term memory. Red indicates higher attention to short-term memory. Temporally consistent regions attend more to long-term memory, while rapidly changing regions attend more to short-term memory.}
    \label{fig:attention_analysis}
\end{figure*}

\vspace{-3mm}
\paragraph{VLM evaluation.}

\begin{wraptable}{r}{0.5\columnwidth}
\vspace{-12pt}
\centering
\scriptsize
\caption{Visual stability (VLM).}
\label{tab:vlm_eval}
\begin{tabular}{l c}
\toprule
Method & Stability \\
\midrule
Self-Forcing~\cite{selfforcing}       & 1.55 \\
Rolling Forcing~\cite{rollingforcing} & 3.40 \\
\midrule
\rowcolor{groupbg}
\multicolumn{2}{l}{\textit{Results on LongLive}~\cite{longlive}} \\
LongLive~\cite{longlive}             & 4.10 \\
Deep Forcing~\cite{deepforcing}       & 3.90 \\
$\infty$-RoPE~\cite{infinityrope}     & 4.05 \\
\textbf{MemRoPE (Ours)}              & \textbf{4.15} \\
\bottomrule
\end{tabular}
\vspace{-10pt}
\end{wraptable}

We further assess long-horizon visual stability using the advanced multimodal vision-language model Gemini 3.1-Pro~\cite{gemini31pro}. Following the protocol of Self-Forcing++~\cite{selfforcingpp} and Deep Forcing~\cite{deepforcing}, we prompt the VLM to score each 120-second generated video in terms of exposure stability and degradation. The prompt and corresponding examples can be found in the supplementary material. As shown in \cref{tab:vlm_eval}, MemRoPE achieves the highest stability scores, consistent with both the automated metrics and the user study.

% \begin{table}[t]
% \centering
% \caption{Ablation on memory components. While using only long-term memory ($\mu_L$) achieves the highest Subject Consistency, combining both memories yields the highest scores in Aesthetic Quality, Imaging Quality, and the overall average. }
% \label{tab:memory_ablation}
% \scriptsize
% \setlength{\tabcolsep}{4pt}
% \begin{tabular}{cc ccc >{\columncolor[gray]{0.9}}c}
% \toprule
% \makecell{Long-term\\Memory ($\mu_L$)} & \makecell{Short-term\\Memory ($\mu_S$)} & \makecell{Aesthetic\\Quality} $\uparrow$ & \makecell{Imaging\\Quality} $\uparrow$ & \makecell{Subject\\Consistency} $\uparrow$ & Avg $\uparrow$ \\
% \midrule
%            &            & 0.5809 & 0.6529 & 0.9740 & 0.8548 \\
% \checkmark &            & 0.5869 & 0.6631 & \textbf{0.9762} & 0.8584 \\
%            & \checkmark & 0.5873 & 0.6663 & 0.9761 & 0.8595 \\
% \checkmark & \checkmark & \textbf{0.5876} & \textbf{0.6709} & 0.9761 & \textbf{0.8597} \\
% \bottomrule
% \end{tabular}
% \end{table}

% Reuses: \first, \second definitions from main table

\begin{table}[t]
\centering
\caption{\textbf{Ablation on memory components.} While using only long-term memory ($\mu_L$) achieves the highest Subject Consistency, combining both memories yields the highest scores in overall average.}
\label{tab:memory_ablation}
\scriptsize
\setlength{\tabcolsep}{4pt}
\resizebox{\columnwidth}{!}{%
\begin{tabular}{cc !{\vrule width 0.4pt} ccc !{\vrule width 0.4pt} c}
\toprule
\makecell{Long-term\\Memory ($\mu_L$)} & \makecell{Short-term\\Memory ($\mu_S$)} & \makecell{Aesthetic\\Quality} $\uparrow$ & \makecell{Imaging\\Quality} $\uparrow$ & \makecell{Subject\\Consistency} $\uparrow$ & Avg $\uparrow$ \\
\midrule
\rowcolor{groupbg}
\multicolumn{6}{l}{\textit{Results on LongLive}~\cite{longlive}} \\
           &            & 58.09 & 65.29 & 97.40 & 85.48 \\
\checkmark &            & \second{58.69} & \second{66.31} & \first{97.62} & \second{85.84} \\
           & \checkmark & 58.73 & 66.63 & \second{97.61} & 85.95 \\
\checkmark & \checkmark & \first{58.76} & \first{67.09} & \second{97.61} & \first{85.97} \\
\bottomrule
\end{tabular}
}%
\end{table}

\vspace{-5mm}

\paragraph{Memory component ablation.}
\label{sec:ablations}
\cref{tab:memory_ablation} isolates the contribution of each memory stream.
Any memory configuration improves over the no-memory baseline, with the largest gain observed in Imaging Quality (+1.8 points).
Combining both streams yields the highest average and the best Imaging Quality among all configurations, suggesting that the two streams capture complementary information: $\mu_L$ summarizes the stable aspects of the scene while $\mu_S$ reflects recent changes.
\cref{fig:attention_analysis} supports this: temporally consistent regions attend more to long-term memory, while rapidly changing regions rely more on short-term memory.

\begin{figure}[t]
    \centering
    \includegraphics[width=\columnwidth]{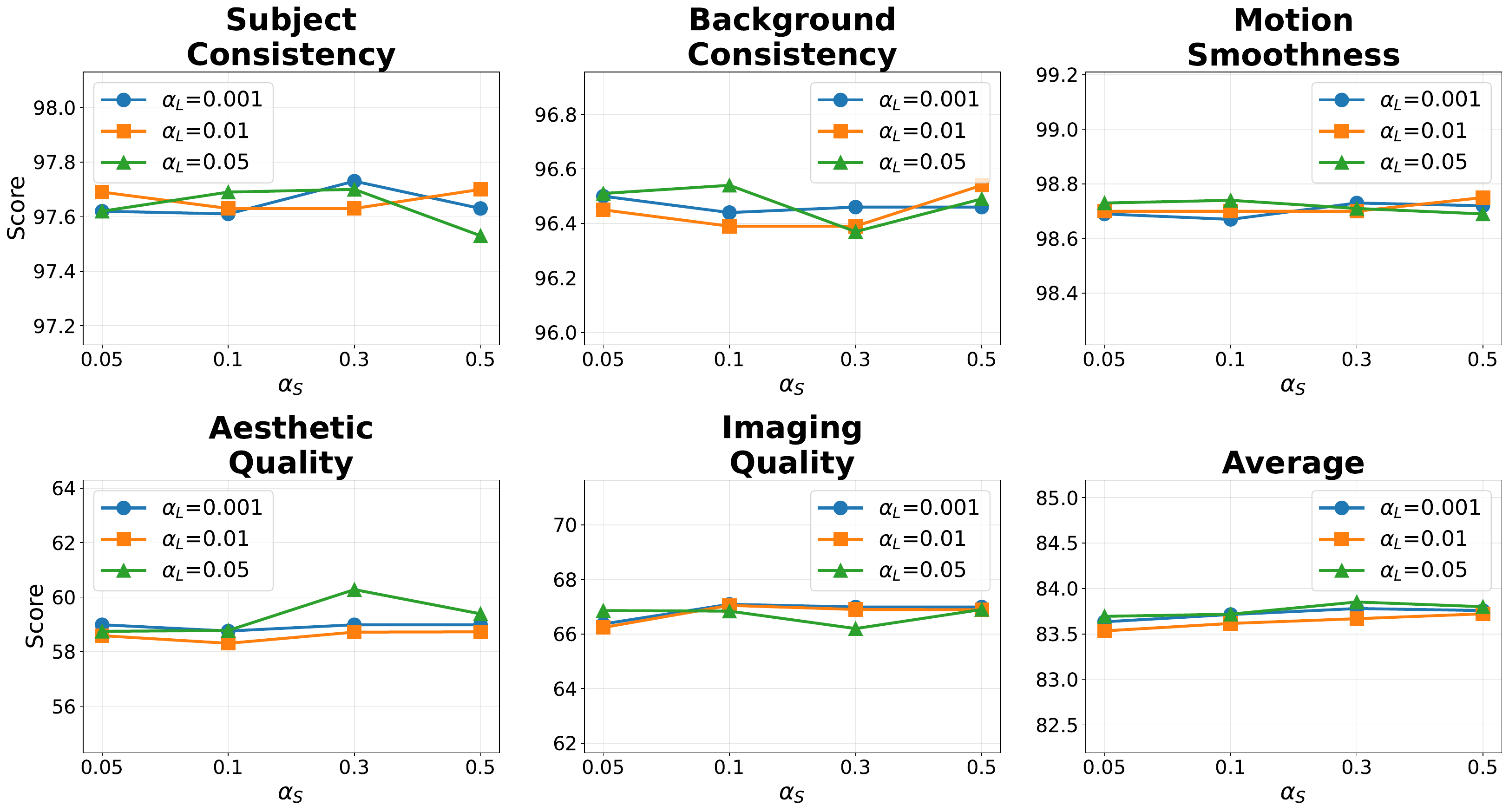}
    \caption{\textbf{Sensitivity to EMA decay rates.} All metrics remain stable across 12 combinations of $\alpha_L$ and $\alpha_S$, with the average score varying by less than 0.7.}
    \label{fig:ema_ablation}
\end{figure}

\vspace{-4mm}
\paragraph{EMA decay rate sensitivity.}
MemRoPE introduces two hyperparameters: the long-term decay $\alpha_L$ and the short-term decay $\alpha_S$. \cref{fig:ema_ablation} plots five VBench metrics across all 12 combinations of $\alpha_L \in \{0.001, 0.01, 0.05\}$ together with $\alpha_S \in \{0.05, 0.1, 0.3, 0.5\}$. The average score varies by less than 0.7 across the entire grid, indicating that MemRoPE is robust to the choice of EMA hyperparameters. We use $\alpha_L{=}0.01$ and $\alpha_S{=}0.1$ for all other experiments.

\vspace{-4mm}

\paragraph{EMA vs.\ attention-based cache management.}
Deep Forcing's participative compression~\cite{deepforcing} is the only other training-free method that goes beyond static sinks for cache management.
As shown in \cref{fig:pc_failure}(c,d), its discrete token selection produces a cache that rarely updates yet causes abrupt visual shifts when it does, whereas our continuous EMA aggregation absorbs every evicted frame smoothly, maintaining stable transitions throughout generation.

% \vspace{-5mm}
% \paragraph{Efficiency.}
% Online RoPE applies elementwise rotations to all cached keys at each attention step, adding less than 0.01\% overhead relative to attention computation.
% EMA updates occur only at chunk eviction and are negligible.
% % MemRoPE thus runs at the same $\sim$20.7 FPS as LongLive~\cite{longlive} on a single H100.

\label{sec:ablation}

\vspace{-1mm}
\section{Conclusion}
\vspace{-1mm}

We present MemRoPE, a training-free framework for long video generation from autoregressive diffusion models. Memory Tokens smoothly preserve evolving context into dual EMA streams, while Online RoPE Indexing makes this well-defined by applying relative positional indices at attention time, enabling unbounded generation within a fixed-size cache. Comprehensive experiments consistently validate the effectiveness of MemRoPE across all tested durations and evaluation protocols. Our results suggest that the key to long-horizon coherence lies not in retaining more frames, but in remembering them better.

\paragraph{Limitations.}
As a training-free method built on a frozen checkpoint, MemRoPE's per-frame quality is bounded by the base model.
The EMA aggregation is also lossy by design, which may limit precise recall of distant content.
Incorporating learned memory compression could address this in future work.

{
    \small
    \bibliographystyle{ieeenat_fullname}
    \bibliography{main}
}

\clearpage
% % Reset layout counters to differentiate from main text
% \setcounter{section}{0}
% \renewcommand{\thesection}{\Alph{section}}
% \setcounter{figure}{0}
% \renewcommand{\thefigure}{S\arabic{figure}}
% \setcounter{table}{0}
% \renewcommand{\thetable}{S\arabic{table}}
% \setcounter{equation}{0}
% \renewcommand{\theequation}{S\arabic{equation}}
% \clearpage
% \newpage
% \appendix

% Create a new title for the supplementary material
% \begin{center}
%     \Large \textbf{Supplementary Material for: \\[0.5em] MemRoPE: Training-Free Infinite Video Generation via Evolving Memory Tokens}
% \end{center}

\appendix
\begin{center}
    \Large \textbf{Appendix}
\end{center}

\noindent This appendix is organized as follows:
\begin{itemize}
    \item Quantitative results on shorter videos (Appendix~\ref{sec:quant_short})
    \item Ablation on position-free caching (Appendix~\ref{sec:roped_ema})
    \item VBench-Long score trends across durations (Appendix~\ref{sec:duration_trend})
    \item Additional qualitative comparisons (Appendix~\ref{sec:qual})
    \item Inference latency measurements (Appendix~\ref{sec:latency})
    \item User study and VLM evaluation details (Appendix~\ref{sec:userstudy})
\end{itemize}

\section{Quantitative Results on Shorter Videos}
\label{sec:quant_short}

The main paper reports VBench-Long~\cite{vbench} results at 120, 240, 480 seconds and 1 hour, where cumulative degradation makes the differences between methods most pronounced. Here we provide complete results at 30 and 60 seconds to demonstrate that MemRoPE is effective even at shorter durations where degradation is minimal.

In addition to the Self-Forcing~\cite{selfforcing} and LongLive~\cite{longlive} base comparisons presented in the main paper, we include additional baselines for reference. CausVid~\cite{causvid} shares the same base model and 5-second training horizon as Self-Forcing, but suffers from a train-inference distribution mismatch---it optimizes against Diffusion Forcing~\cite{diffusionforcing} outputs rather than the actual inference-time distribution---leading to over-exposure artifacts that worsen at longer durations. We adopt Self-Forcing as a base model because it resolves this mismatch by simulating the actual inference process during training, providing a stronger foundation for long video generation. Rolling Forcing~\cite{rollingforcing} performs additional distillation training and adopts a fundamentally different inference paradigm based on rolling-window joint denoising, making it incompatible as a base model for our plug-in method. NOVA~\cite{nova}, MAGI-1~\cite{magi1}, and SkyReels-V2~\cite{skyreels} use entirely different architectures. These were omitted from the main tables to keep the comparison focused on training-free methods under controlled base-model settings, but are included here for completeness.

MemRoPE achieves the highest average on both Self-Forcing and LongLive bases at both durations. Notably, even at 30 seconds where degradation has not yet accumulated significantly, MemRoPE already outperforms all competing methods within each base group, indicating that our memory mechanism improves generation quality from the outset rather than only compensating for long-horizon drift.

\begin{table*}[t]
\centering
\caption{\textbf{Quantitative comparison on short video generation (30 and 60 seconds).} We report VBench-Long~\cite{vbench} metrics. Within each base model group, best results are {bold with darker background} and second best are with lighter background. $\Delta$ denotes the average improvement over the respective base model. $^*$Results from~\cite{infinityrope}.}
\label{tab:short}
\resizebox{\textwidth}{!}{%
\begin{tabular}{l cccccc !{\vrule width 0.8pt} c c}
\toprule
Method & \makecell{Aesthetic\\Quality} $\uparrow$ & \makecell{Background\\Consistency} $\uparrow$ & \makecell{Imaging\\Quality} $\uparrow$ & \makecell{Motion\\Smoothness} $\uparrow$ & \makecell{Subject\\Consistency} $\uparrow$ & \makecell{Temporal\\Flickering} $\uparrow$ & Average $\uparrow$ & $\Delta$ \\
\midrule
\multicolumn{9}{c}{\textit{30 seconds}} \\
\midrule
CausVid~\cite{causvid}                     & 57.25 & 96.77 & 65.16 & 98.08 & 97.87 & 96.86 & 85.33 & -- \\
Rolling Forcing~\cite{rollingforcing}      & 58.45 & 95.77 & 69.35 & 98.23 & 96.78 & 97.13 & 85.95 & -- \\
\midrule
\rowcolor{groupbg}
\multicolumn{9}{l}{\textit{Results on Self-Forcing}~\cite{selfforcing}} \\
Self-Forcing~\cite{selfforcing}            & \second{58.28} & 95.14 & \second{66.92} & \second{98.02} & 95.50 & \second{96.87} & \second{85.12} & -- \\
Deep Forcing~\cite{deepforcing}            & 57.83 & 94.86 & 66.29 & 97.53 & 94.98 & 95.71 & 84.53 & $-$0.59 \\
$\infty$-RoPE~\cite{infinityrope}          & 57.13 & \first{95.81} & 65.10 & \first{98.39} & \first{96.37} & \first{97.18} & 85.00 & $-$0.12 \\
\textbf{MemRoPE (Ours)}                    & \first{58.48} & \second{95.59} & \first{68.73} & 98.01 & \second{96.31} & 96.41 & \first{85.59} & \first{$+$0.47} \\
\addlinespace[2pt]
\rowcolor{groupbg}
\multicolumn{9}{l}{\textit{Results on LongLive}~\cite{longlive}} \\
LongLive~\cite{longlive}                   & 58.57 & 96.22 & 67.85 & 98.65 & 97.21 & 97.32 & 85.97 & -- \\
Deep Forcing~\cite{deepforcing}            & \first{59.87} & 96.22 & \first{69.41} & 98.59 & 97.05 & 97.33 & \second{86.41} & \second{$+$0.44} \\
$\infty$-RoPE~\cite{infinityrope}          & \second{59.32} & \first{96.40} & 66.68 & \first{98.94} & \second{97.32} & \first{97.93} & 86.10 & $+$0.13 \\
\textbf{MemRoPE (Ours)}                    & 59.31 & \second{96.34} & \second{69.27} & \second{98.68} & \first{97.48} & \second{97.40} & \first{86.42} & \first{$+$0.45} \\
\midrule
%%%%%%%%%%%%%%%%%%%%%%%%%% 60s %%%%%%%%%%%%%%%%%%%%%%%%%%
\multicolumn{9}{c}{\textit{60 seconds}} \\
\midrule
NOVA$^*$~\cite{nova}                           & 47.53 & 88.06 & 44.97 & 98.94 & 77.50 & 98.27 & 75.88 & -- \\
MAGI-1$^*$~\cite{magi1}                        & 52.10 & 87.76 & 54.54 & 99.26 & 79.46 & 98.48 & 78.60 & -- \\
SkyReels-V2$^*$~\cite{skyreels}              & 57.64 & 89.95 & 66.67 & 98.67 & 84.99 & 97.60 & 82.59 & -- \\

CausVid~\cite{causvid}                     & 57.06 & 96.94 & 64.84 & 98.16 & 98.03 & 97.05 & 85.35 & -- \\
Rolling Forcing~\cite{rollingforcing}      & 58.34 & 96.21 & 69.41 & 98.52 & 97.26 & 97.47 & 86.20 & -- \\
\midrule
\rowcolor{groupbg}
\multicolumn{9}{l}{\textit{Results on Self-Forcing}~\cite{selfforcing}} \\
Self-Forcing~\cite{selfforcing}            & 56.96 & 95.08 & \second{66.36} & 97.35 & 95.15 & \second{96.56} & \second{84.57} & -- \\
Deep Forcing~\cite{deepforcing}            & \second{57.30} & 94.59 & 65.67 & 97.22 & 94.36 & 95.38 & 84.09 & $-$0.48 \\
$\infty$-RoPE~\cite{infinityrope}          & 54.97 & \first{95.69} & 62.81 & \first{98.39} & \second{96.26} & \first{97.37} & 84.25 & $-$0.32 \\
\textbf{MemRoPE (Ours)}                    & \first{57.77} & \second{95.58} & \first{68.54} & \second{97.93} & \first{96.29} & 96.35 & \first{85.41} & \first{$+$0.84} \\
\addlinespace[2pt]
\rowcolor{groupbg}
\multicolumn{9}{l}{\textit{Results on LongLive}~\cite{longlive}} \\
LongLive~\cite{longlive}                   & 57.93 & 96.10 & 67.41 & 98.57 & 97.10 & 97.15 & 85.71 & -- \\
Deep Forcing~\cite{deepforcing}            & \first{59.63} & 96.16 & \second{68.73} & 98.52 & 96.88 & 97.33 & \second{86.21} & \second{$+$0.50} \\
$\infty$-RoPE~\cite{infinityrope}          & 58.52 & \first{96.33} & 65.63 & \first{98.95} & \second{97.27} & \first{97.95} & 85.78 & $+$0.07 \\
\textbf{MemRoPE (Ours)}                    & \second{58.76} & \second{96.27} & \first{69.01} & \second{98.67} & \first{97.41} & \second{97.34} & \first{86.24} & \first{$+$0.53} \\
\bottomrule
\end{tabular}%
}
\end{table*}

\section{Ablation on Position-Free Caching}
\label{sec:roped_ema}

As discussed in Sec.~4.2 of the main paper, RoPE does not distribute over addition (Eq.~(5)): averaging keys that carry different rotary phases produces representations that no longer correspond to any valid temporal position. A natural question is whether this theoretical concern has a measurable impact in practice. To answer this, we compare MemRoPE (position-free EMA) against an EMA variant that retains RoPE embeddings during aggregation (Aggregation w/ RoPE).

As shown in Tab.~\ref{tab:ablation_ema}, both methods improve over their respective base models, confirming that temporal aggregation is broadly beneficial. However, MemRoPE consistently achieves a higher average improvement ($\Delta$) on both Self-Forcing ($+$0.84 vs.\ $+$0.71) and LongLive ($+$0.53 vs.\ $+$0.37). While Aggregation w/ RoPE scores higher on certain individual metrics, its overall coherence is lower due to the conflicting rotary phases discussed in Eq.~(5). This validates our design choice: decoupling positional information from cached keys before temporal aggregation is essential for well-defined memory compression.

\begin{table*}[t]
\centering
\caption{\textbf{Ablation on position-free caching.} Comparison between MemRoPE (position-free EMA) and EMA with RoPE-rotated keys at 60 seconds. $\Delta$ denotes the average improvement over the respective base model.}
\label{tab:ablation_ema}
\resizebox{\textwidth}{!}{%
\begin{tabular}{l cccccc  !{\vrule width 0.8pt} c  c}
\toprule
Method & \makecell{Aesthetic\\Quality} $\uparrow$ & \makecell{Subject\\Consistency} $\uparrow$ & \makecell{Imaging\\Quality} $\uparrow$ & \makecell{Motion\\Smoothness} $\uparrow$ & \makecell{Background\\Consistency} $\uparrow$ & \makecell{Temporal\\Flickering} $\uparrow$ & Average $\uparrow$ & $\Delta$ \\
\midrule
\rowcolor{groupbg}
\multicolumn{9}{l}{\textit{Results on Self-Forcing}~\cite{selfforcing}} \\
Self-Forcing~\cite{selfforcing}   & \second{56.96} & 95.08 & 66.36 & 97.35 & 95.15 & \first{96.66} & 84.57 & -- \\
Aggregation w/ RoPE                & 56.55 & \first{96.44} & \second{68.25} & \first{98.19} & \second{95.60} & \first{96.66} & \second{85.28} & \second{$+$0.71} \\
\textbf{MemRoPE (Ours)}           & \first{57.77} & \second{95.58} & \first{68.54} & \second{97.93} & \first{96.29} & 96.35 & \first{85.41} & \first{$+$0.84} \\
\addlinespace[4pt]
\rowcolor{groupbg}
\multicolumn{9}{l}{\textit{Results on LongLive}~\cite{longlive}} \\
LongLive~\cite{longlive}          & 57.93 & 96.10 & 67.41 & \second{98.57} & \second{97.10} & 97.15 & 85.71 & -- \\
Aggregation w/ RoPE                & \second{58.50} & \first{97.35} & \second{68.45} & \first{98.67} & 96.18 & \second{97.33} & \second{86.08} & \second{$+$0.37} \\
\textbf{MemRoPE (Ours)}           & \first{58.76} & \second{96.27} & \first{69.01} & \first{98.67} & \first{97.41} & \first{97.34} & \first{86.24} & \first{$+$0.53} \\
\bottomrule
\end{tabular}%
}
\end{table*}

\section{VBench-Long Score Across Durations}
\label{sec:duration_trend}

\begin{figure}[h]
    \centering
    \includegraphics[width=\columnwidth]{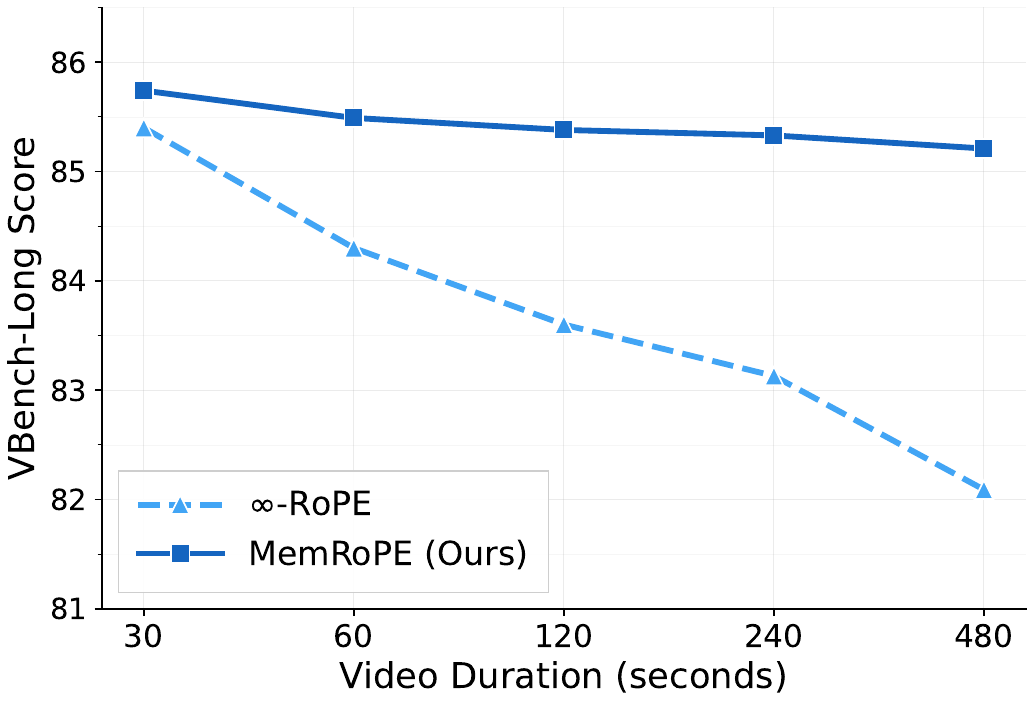}
    \caption{\textbf{VBench-Long average score vs.\ video duration on Self-Forcing~\cite{selfforcing}.} MemRoPE maintains higher scores across all durations, with the gap widening as videos grow longer.}
    \label{fig:duration_trend}
\end{figure}

We collect the VBench-Long averages of MemRoPE and $\infty$-RoPE~\cite{infinityrope} on Self-Forcing~\cite{selfforcing} across five durations: 30, 60, 120, 240, and 480 seconds. To ensure a consistent comparison, we report results using the 20-prompt subset used for the 480-second evaluation across all durations. As shown in Fig.~\ref{fig:duration_trend}, both methods degrade as duration increases, but MemRoPE exhibits a significantly flatter slope. The performance gap widens monotonically as duration increases, demonstrating that the benefits of our evolving memory tokens compound over time. This scaling behavior confirms that MemRoPE's dual-stream EMA effectively preserves long-range context that $\infty$-RoPE fully discards.

\section{Qualitative Comparisons}
\label{sec:qual}

\begin{figure*}[h]
    \centering
    \includegraphics[width=1.0\textwidth]{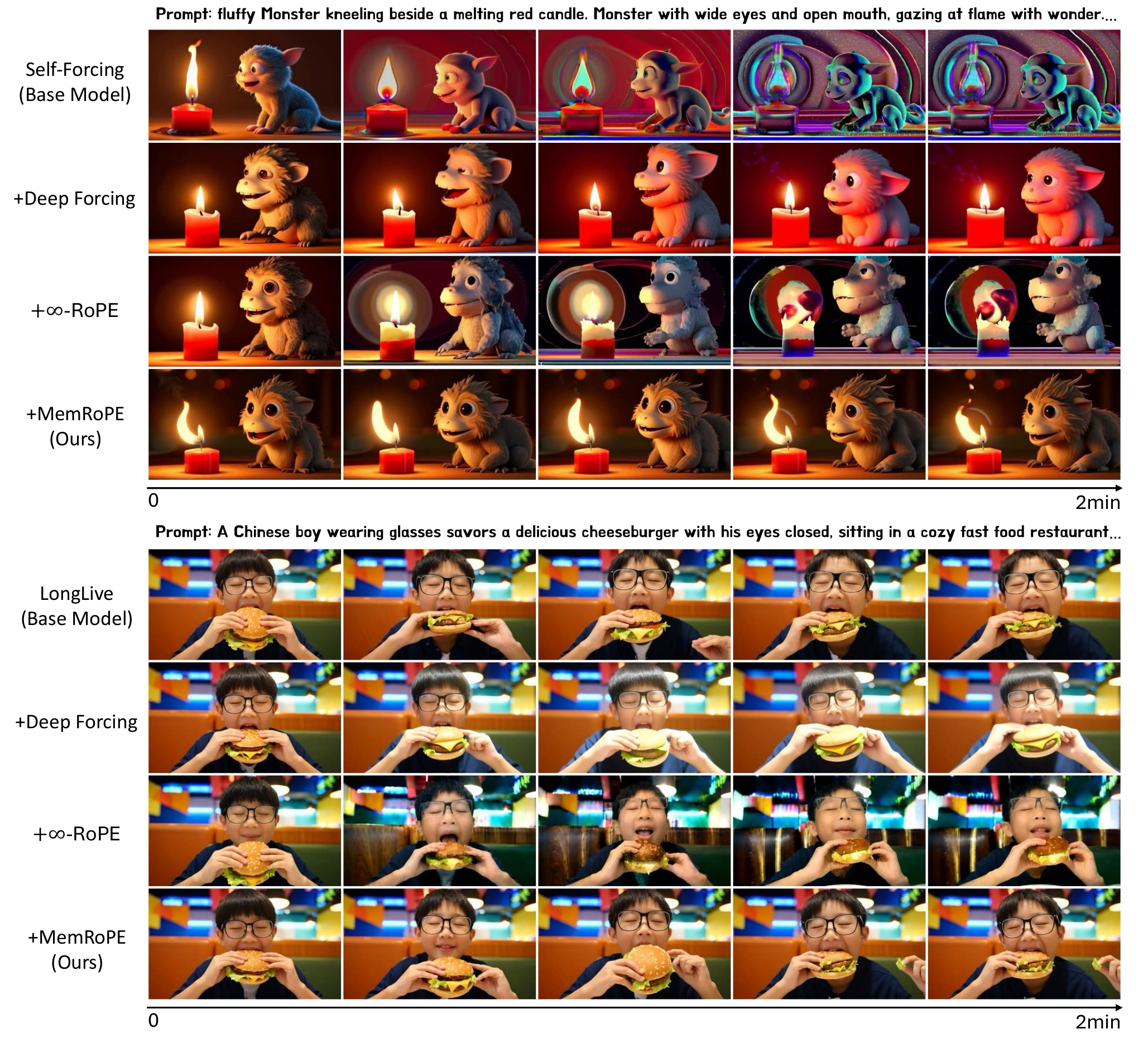}
    \caption{\textbf{Qualitative comparison at 2-minute generation on Self-Forcing~\cite{selfforcing} and LongLive~\cite{longlive}.} MemRoPE preserves subject identity and visual fidelity more consistently than existing methods across both prompts.}
    \label{fig:qual_2min}
\end{figure*}

\begin{figure*}[h]
    \centering
    \includegraphics[width=\textwidth]{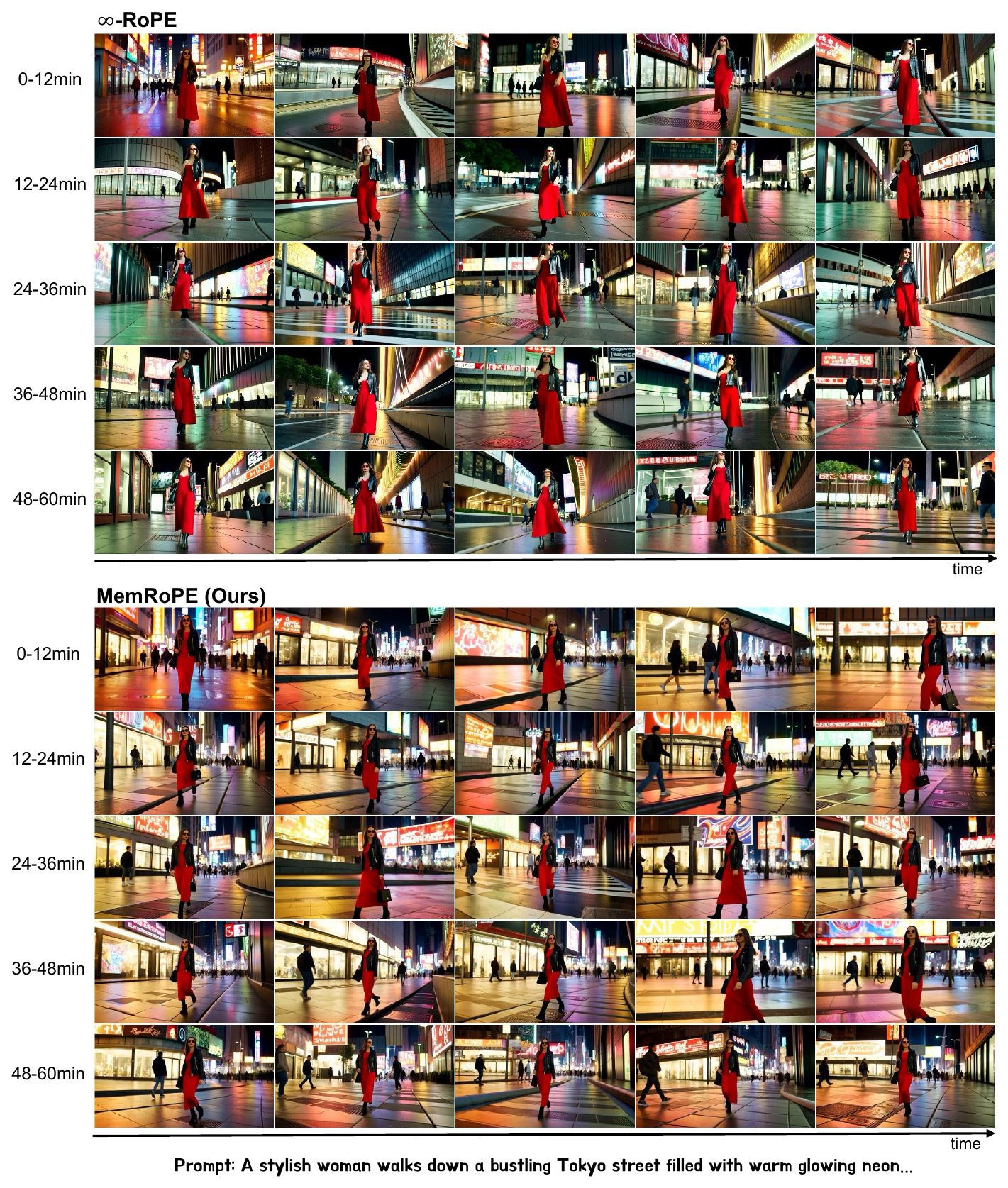}
    \caption{\textbf{1-hour generation comparison on LongLive~\cite{longlive}.} Top: $\infty$-RoPE~\cite{infinityrope}. Bottom: MemRoPE (Ours). Frames sampled at 12-minute intervals. MemRoPE maintains more consistent subject appearance and background color tone over the full hour.}
    \label{fig:1hour}
\end{figure*}

\noindent\textbf{2-Minute Generation.}
Fig.~\ref{fig:qual_2min} compares MemRoPE against Self-Forcing~\cite{selfforcing}, Deep Forcing~\cite{deepforcing}, and $\infty$-RoPE~\cite{infinityrope} on two prompts at 2-minute duration. On Self-Forcing, the base model degrades severely by the end of generation. Deep Forcing preserves the general structure but the main subject undergoes noticeable shape changes over time. $\infty$-RoPE loses subject consistency entirely, with colors collapsing across frames. On LongLive, the base model exhibits gradual subject appearance drift, Deep Forcing shifts the overall color tone, and $\infty$-RoPE suffers from both appearance changes and color corruption. MemRoPE maintains consistent subject identity and visual fidelity throughout on both base models.

\vspace{3mm}
\noindent\textbf{1-Hour Generation.}
Fig.~\ref{fig:1hour} compares frames sampled at 12-minute intervals from hour-long videos generated by $\infty$-RoPE~\cite{infinityrope} and MemRoPE on LongLive~\cite{longlive}. Both methods maintain the general scene structure, but $\infty$-RoPE shows gradual inconsistency in the subject's facial features and hairstyle, along with increasingly saturated background colors in later segments. MemRoPE preserves more stable subject appearance and background color tone throughout the full hour.

\begin{figure*}[t]
    \centering
    \includegraphics[width=1.0\textwidth]{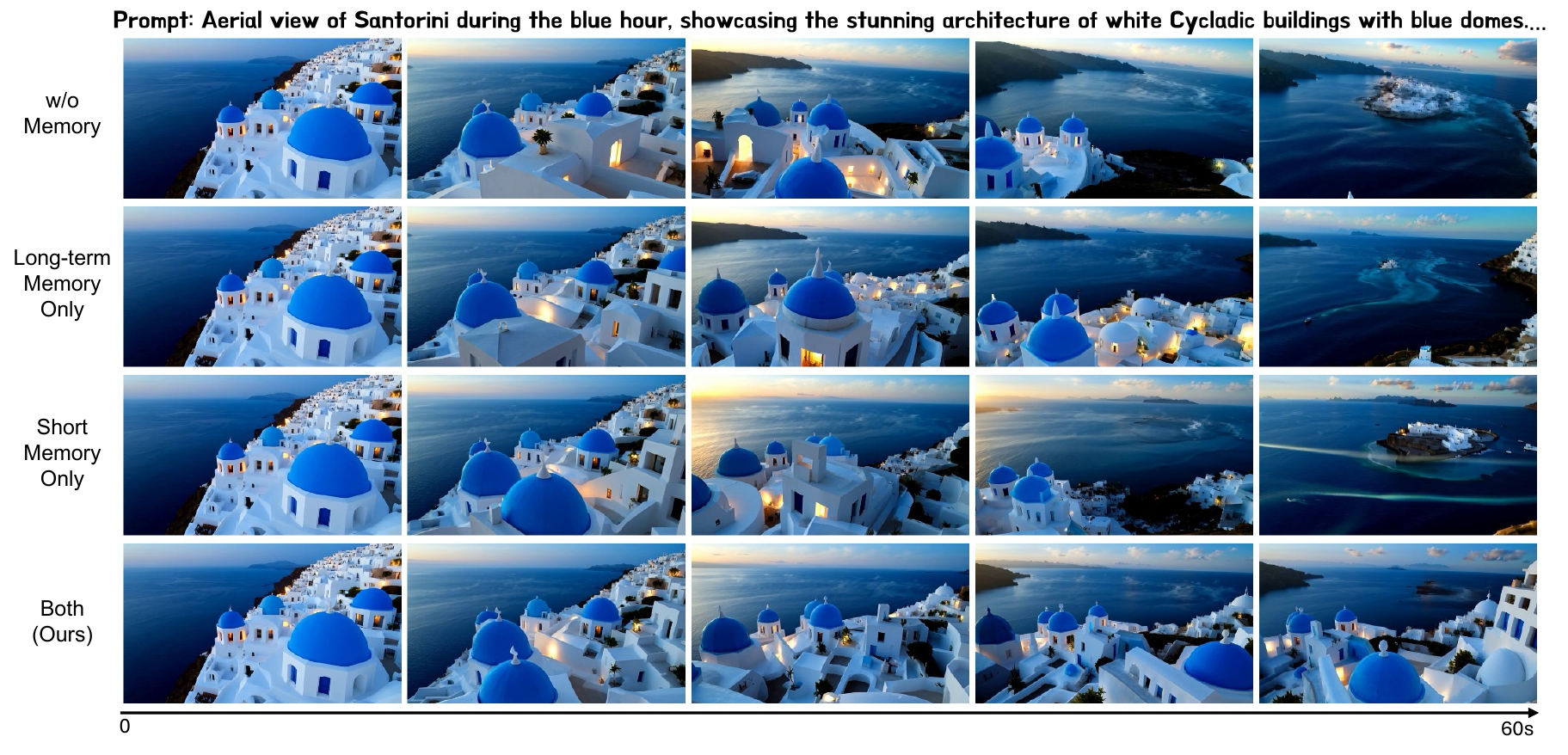}
    \caption{\textbf{Qualitative memory component ablation on LongLive~\cite{longlive}.} From top to bottom: no memory, long-term memory only, short-term memory only, and both (ours). Only the full dual-stream configuration preserves background structures and subject appearance throughout.}
    \label{fig:mem_ablation}
\end{figure*}

% \begin{figure}[t]
%     \centering
%     \includegraphics[width=1.0\textwidth]{fig/supp_1hour_infrope.pdf}
%     \caption{\textbf{1-hour generation with $\infty$-RoPE~\cite{infinityrope} on LongLive~\cite{longlive}.} Frames sampled at 10-minute intervals. Color drift and background inconsistency become apparent.}
%     \label{fig:1hour_infrope}
% \end{figure}

% \begin{figure}[t]
%     \centering
%     \includegraphics[width=1.0\textwidth]{fig/supp_1hour_ours.pdf}
%     \caption{\textbf{1-hour generation with MemRoPE (Ours) on LongLive~\cite{longlive}.} Frames sampled at 10-minute intervals. Subject identity and scene fidelity remain stable throughout the full hour.}
%     \label{fig:1hour_ours}
% \end{figure}

\vspace{3mm}
\noindent\textbf{Memory Component Ablation.}
Fig.~\ref{fig:mem_ablation} visualizes the effect of each memory component on LongLive~\cite{longlive} at 60 seconds. Without any memory, with long-term memory only, and with short-term memory only, background structures such as buildings gradually disappear as generation progresses. Only when both streams are combined does the full method preserve the background architecture and subject appearance throughout the sequence, suggesting that both memory scales are jointly necessary for maintaining scene integrity.

\section{Inference Latency}
\label{sec:latency}

We measure inference speed on a single NVIDIA A6000 GPU after generating one initial chunk to warm up the model and GPU state. As shown in Tab.~\ref{tab:latency}, MemRoPE adds negligible overhead at the same cache size (LongLive, both $C=12$). On Self-Forcing, MemRoPE is \emph{faster} because its compact three-tier cache ($C=12$) replaces the original sliding window ($C=21$), reducing the number of tokens attended to during each denoising step.

\begin{table}[h]
\centering
\small
\caption{\textbf{Inference speed.} Video FPS measured on a single A6000 GPU.}
\label{tab:latency}
\begin{tabular}{l c}
\toprule
Method & Video FPS \\
\midrule
Self-Forcing~\cite{selfforcing} ($C=21$) & 3.631 \\
+ MemRoPE ($C=12$) & 5.115 \\
\midrule
LongLive~\cite{longlive} ($C=12$) & 4.409 \\
+ MemRoPE ($C=12$) & 4.376 \\
\bottomrule
\end{tabular}
\end{table}

\section{VLM Evaluation and User Study  Details}
\label{sec:userstudy}

% \vspace{-10mm}
\noindent\textbf{VLM-Based Evaluation.}
In addition to the user study, we employ Gemini 3.1 Pro~\cite{gemini31pro} as a VLM judge to evaluate generated videos on a 5-point exposure stability scale, following Self-Forcing++~\cite{selfforcingpp}. The evaluation prompt is as follows:

\begin{tcolorbox}[colback=gray!5, colframe=gray!50, title=VLM Evaluation Prompt, fonttitle=\bfseries\small]
\small
You are tasked with rating the exposure stability of a video. Assign a score according to the following scale:

\medskip
\textbf{0:} Catastrophic Exposure. Nearly the entire frame is either blown out (pure white) or crushed (pure black), rendering the scene unreadable.

\textbf{1:} Severe Exposure Failure. Large portions are dominated by over- or under-exposure, substantially impairing visibility.

\textbf{2:} Noticeable Exposure Problems. Persistent clipping in highlights or shadows; significant areas lose detail.

\textbf{3:} Moderate Exposure Issues. Over-exposed highlights or under-exposed shadows occur but are limited in extent or duration.

\textbf{4:} Minor Exposure Flaws. Small regions are occasionally too bright or too dark, but do not meaningfully disrupt visibility.

\textbf{5:} Well-Exposed. Balanced lighting across the frame with no distracting over-exposure or darkening.
\end{tcolorbox}

\begin{figure*}[h!]
    \centering
    \includegraphics[width=0.9\textwidth]{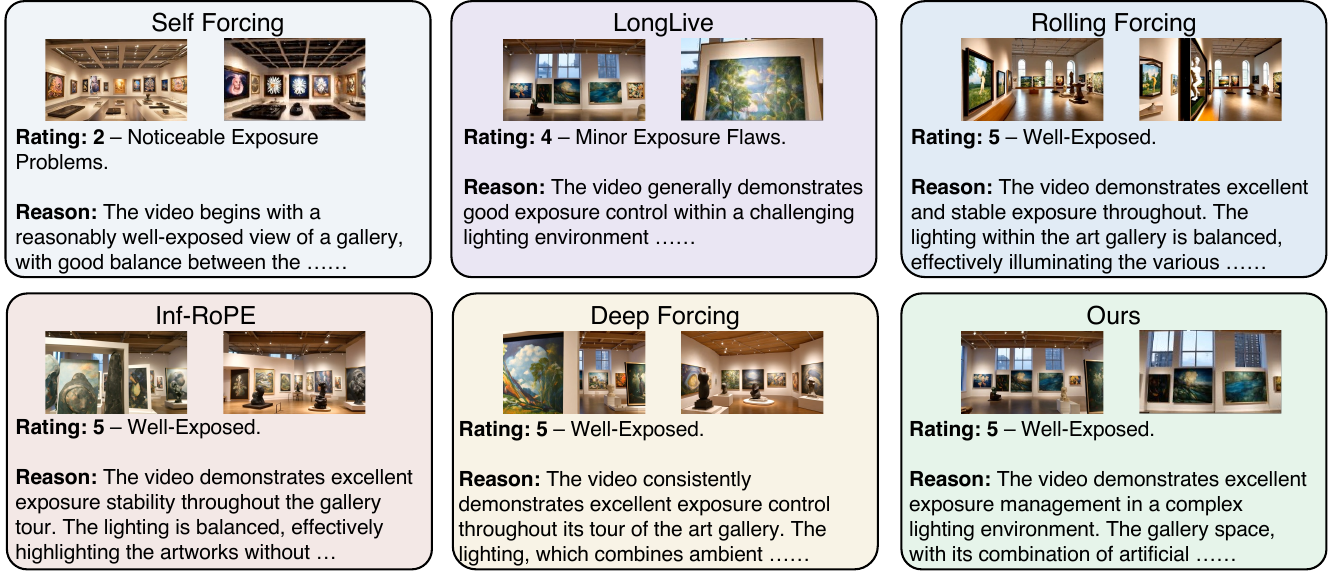}
    \caption{\textbf{Example VLM evaluation.} Gemini 3.1 Pro~\cite{gemini31pro} rates each method on a 5-point scale with detailed reasoning.}
    \label{fig:supfig_vlm_b}
\end{figure*}

\noindent Fig.~\ref{fig:supfig_vlm_b} shows example evaluations.
% \clearpage

\begin{figure*}[h!]
    \centering
    \includegraphics[width=0.8\textwidth]{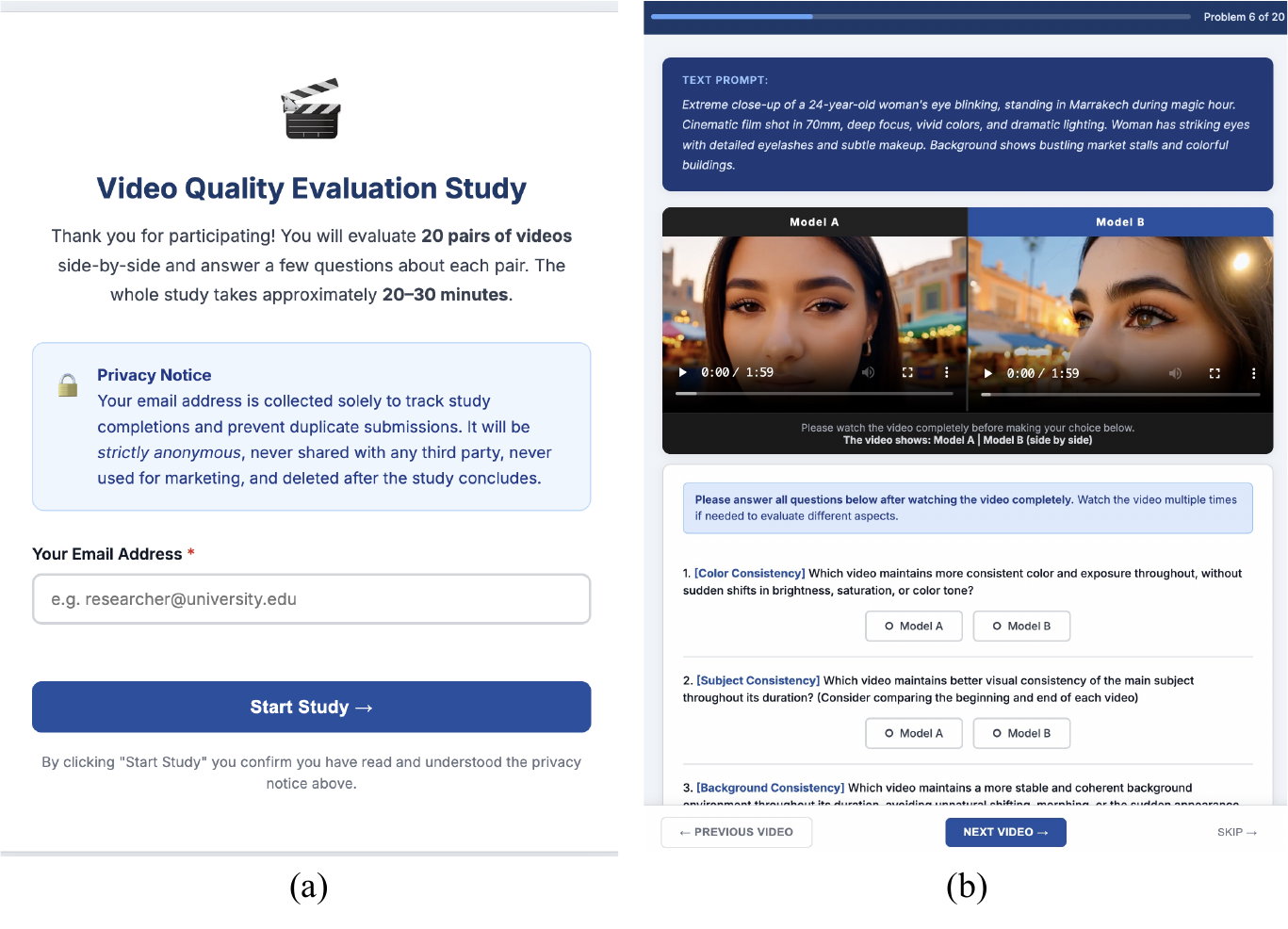}
    \caption{\textbf{User study interface.} (a) Welcome page with instructions. (b) Evaluation page showing side-by-side video comparison with per-dimension preference questions.}
    \label{fig:supfig_usersty}
\end{figure*}

\vspace{3mm}
\noindent\textbf{User Study Interface.}
Fig.~\ref{fig:supfig_usersty} shows our user study interface. Participants evaluate 20 pairs of side-by-side videos generated from the same prompt by two anonymized methods. For each pair, they answer six questions covering Color Consistency, Subject Consistency, Background Consistency, Text Alignment, Motion Smoothness, and Overall Preference. 
% The order of methods is randomized per pair to avoid positional bias.

\end{document}